\documentclass[letterpaper]{article} % DO NOT CHANGE THIS
\usepackage{aaai24}  % DO NOT CHANGE THIS
\usepackage{times}  % DO NOT CHANGE THIS
\usepackage{helvet}  % DO NOT CHANGE THIS
\usepackage{courier}  % DO NOT CHANGE THIS
\usepackage[hyphens]{url}  % DO NOT CHANGE THIS
\usepackage{graphicx} % DO NOT CHANGE THIS
\usepackage{natbib}  % DO NOT CHANGE THIS AND DO NOT ADD ANY OPTIONS TO IT
\usepackage{caption} % DO NOT CHANGE THIS AND DO NOT ADD ANY OPTIONS TO IT
\usepackage{amsmath,amssymb}
\usepackage{algorithm}
\usepackage{algorithmic}
\usepackage{multirow}
\usepackage{colortbl}

\usepackage{newfloat}
\usepackage{listings}
\usepackage{marvosym}
\usepackage{lipsum}
\DeclareCaptionStyle{ruled}{labelfont=normalfont,labelsep=colon,strut=off} % DO NOT CHANGE THIS
\floatstyle{ruled}
\newfloat{listing}{tb}{lst}{}
\floatname{listing}{Listing}

\newcommand\blfootnote[1]{%
  \begingroup
  \renewcommand\thefootnote{}\footnote{#1}%
  \addtocounter{footnote}{-1}%
  \endgroup
}
\title{SeTformer is What You Need for Vision and Language}
\author{
    Pourya Shamsolmoali\textsuperscript{\rm 1},
    Masoumeh Zareapoor\textsuperscript{\rm 2},
    Eric Granger\textsuperscript{\rm 3},
    Michael Felsberg\textsuperscript{\rm 4}
}
\affiliations{
    \textsuperscript{\rm 1}East China Normal University,  
     \textsuperscript{\rm 2}Shanghai Jiaotong University,  
     \textsuperscript{\rm 3}ETS Montreal, 
     \textsuperscript{\rm 4}Link\"oping University\\ 
    \blfootnote{Correspondence to $<$mzarea222@gmail.com$>$}
}

\usepackage{bibentry}
\begin{document}

\maketitle

\begin{abstract}
%%%%%%%%%%%%%%%%%%%%%%%
The dot product self-attention (DPSA) is a fundamental component of transformers. However, scaling them to long sequences, like documents or high-resolution images, becomes prohibitively expensive due to quadratic time and memory complexities arising from the softmax operation. Kernel methods are employed to  simplify computations by approximating softmax but often lead to performance drops compared to softmax attention. We propose SeTformer, a novel transformer, where DPSA is purely replaced by {\bf Se}lf-optimal {\bf T}ransport (SeT) for achieving better performance and computational efficiency. SeT is based on two essential softmax properties: maintaining a non-negative attention matrix and using a nonlinear reweighting mechanism to emphasize important tokens in input sequences. By introducing a kernel cost function for optimal transport, SeTformer effectively satisfies these properties. 
%%%%%%%%%%%%%%%%%%%%%%%%%%%%%%%%%%%%
In particular, with small and base-sized models, SeTformer achieves impressive top-1 accuracies of 84.7\% and 86.2\% on ImageNet-1K. In object detection, SeTformer-base outperforms the FocalNet counterpart by +2.2 mAP, using 38\% fewer parameters and 29\% fewer FLOPs. In semantic segmentation, our base-size model surpasses NAT by +3.5 mIoU with 33\% fewer parameters. SeTformer also achieves state-of-the-art results in language modeling on the GLUE benchmark. These findings highlight SeTformer's applicability in vision and language tasks.

\end{abstract}

%\vspace{-10pt}
%%%%%%%%%%%%%%%%%%%%%%
\section{Introduction}
\begin{figure}[t]
\centering
 \includegraphics[width=8.4cm]{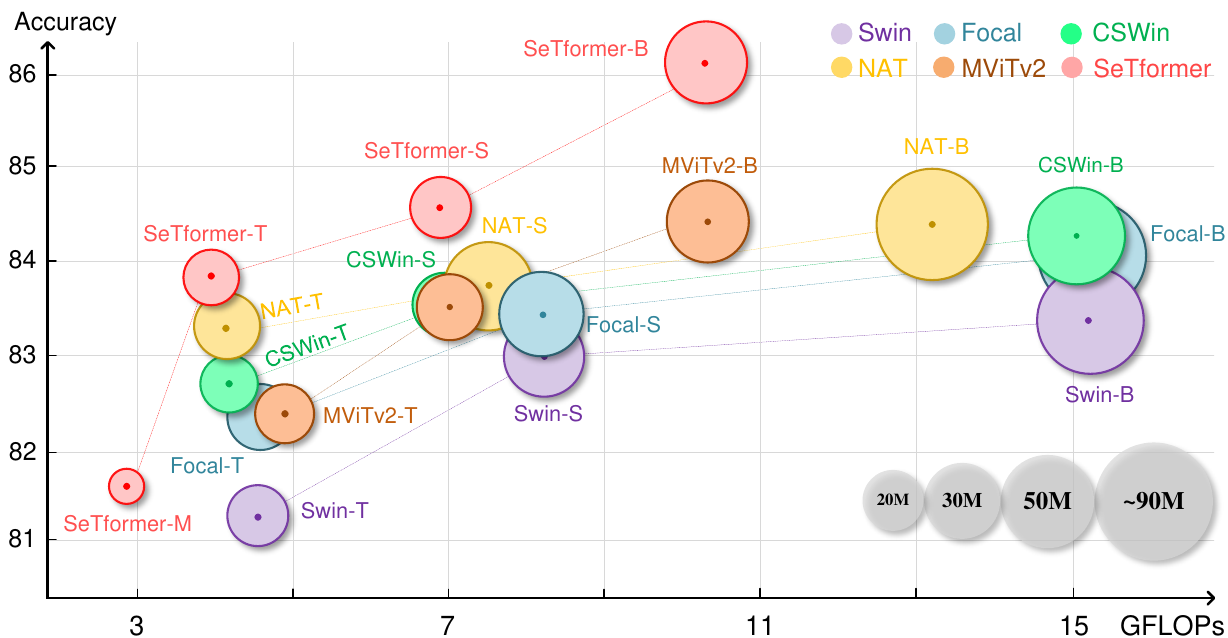}
  \caption{Top-1 classification accuracy vs. FLOPs on the ImageNet-1k, where the bubble size represents the number of parameters. SeTformer improves upon the baseline accuracy yet require fewer parameters and FLOPs.}
\label{fig1}
\end{figure}

\begin{figure}[t]
\centering
 \includegraphics[width=8.3cm]{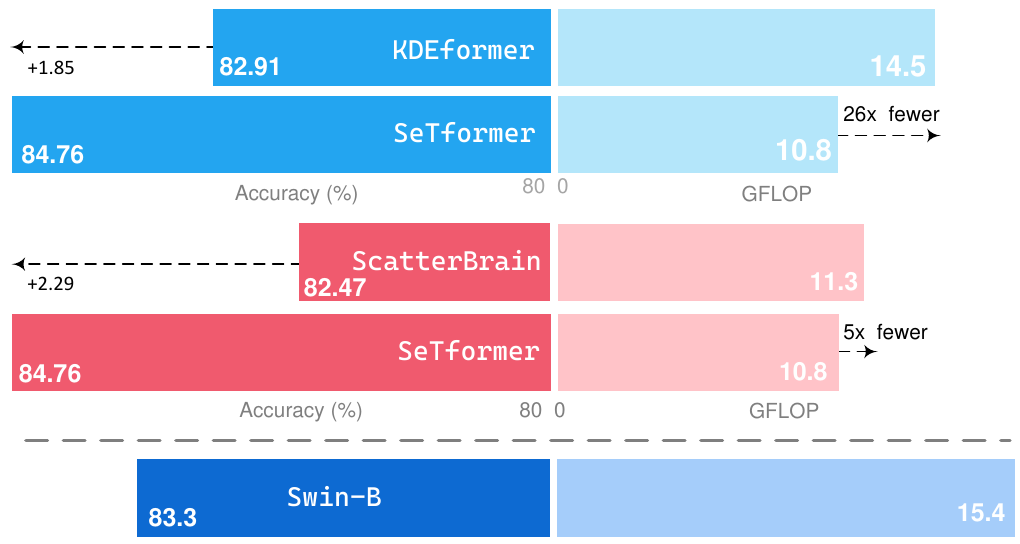}
 %\vspace{-5pt}
 \caption{Top-1 Classification accuracy vs. FLOPs on the ImageNet-1k. We compare SeTformer with other kernel-based transformers, while using Swin-B as the base.}
\label{kernel}
\end{figure}

Transformers \cite{vaswani2017attention}, initially introduced for natural language processing (NLP), have gained significant popularity in computer vision following the groundbreaking work of Vision Transformer (ViT) \cite{dosovitskiy2020image}. Its promise has been demonstrated in various vision tasks, including image classification, object detection, segmentation and beyond \cite{khan2022transformers}. Dot-product self-attention (DPSA) with softmax normalization plays a crucial role in transformers for capturing long-range dependencies. However, the computation of this model leads to quadratic time and memory complexities, making it challenging to train long sequence models. Therefore, research on ``efficient" transformers has gained significant importance in last two years. 
%%%%%%%%%%%%%%%%
Several methods with the aim of increasing the efficiency of transformers  
\cite{touvron2021training, liu2021swin, huang2022lightvit, yang2022focal, liu2022convnet, wang2022pvt,hassani2023neighborhood, grainger2023paca} are proposed. Although these methods have achieved impressive results on ImageNet-1K (an area where ViT has struggled), the increased complexity makes these models slower overall (Figure \ref{fig1}). 
%Although these methods have achieved impressive results  , striking a balance between model accuracy and efficiency remains a challenge (Fig. \ref{fig1}). The increased FLOP counts can be attractive for performance metrics, but it comes at the expense of computational efficiency and model complexity.
%%%%%%%%%%%%%%%%%%%%%%%%%%%%%%%%%%%%%
%Similarly,for language modeling approaches like \cite{zaheer2020big, beltagy2020longformer, tay2020sparse, kitaev2020reformer}  have demonstrated impressive results. 
%%%
%This limitation affects various tasks. For example, in NLP, document-level tasks and language models often involve processing long sequences \cite{raffel2020exploring, roy2021efficient, tay2022efficient}. Similarly, in computer vision, tasks dealing with high-resolution images require transforming them into long sequences of image patches \cite{wu2021cvt, dosovitskiy2020image, touvron2021training} before being processed with transformers. %Therefore, developing an efficient attention mechanism for long sequence modeling that generalizes across domains is crucial. 
%However, an efficient attention mechanism that generalizes well in both language and vision domains is less explored. 
%%%%%%
%%%%
%%%%%%%%
One main bottleneck in attention mechanisms is the softmax operator, which affects the efficiency \cite{zandieh2023kdeformer}. Thus, recent research has proposed kernel-based methods as a way to approximate the attention matrix and improve computational efficiency \cite{choromanski2020rethinking, peng2021random, chowdhury2022learning, choromanski2023learning, zandieh2023kdeformer, reid2023simplex}.
%%%
%However, the softmax operator is considered to be the main bottleneck that hinders efficiency in attention mechanisms. To address this, recent research has proposed kernel-based methods as a way to approximate the attention matrix and improve computational efficiency \cite{tsai2019transformer, choromanski2020rethinking, peng2021random, chowdhury2022learning, choromanski2023learning, zandieh2023kdeformer, reid2023simplex}.
%
%Moreover, the softmax operator is considered to be the main bottleneck to achieve the efficiency, recent research has explored more efficient attention through kernel-based approximation of the attention matrix \cite{tsai2019transformer, choromanski2020rethinking, peng2021random, chowdhury2022learning, koohpayegani2022sima, choromanski2023learning, zandieh2023kdeformer, reid2023simplex}. %Recently, there has been increased interest in constructing dense attention matrices using low-rank kernels instead of softmax to reduce the computation complexity. 
%%%%%
The approximate attention matrix is an unbiased estimate of the original attention matrix, encoding token similarities via a softmax-kernel without explicit construction. 
%%%%%%%%%%%%%%%%%%%%%%%%%%%%%%%
%%%%%%%%%%%%%%%%%%%%%%%%%%
%%%%%%%%%%%%%%%%%%%%%%%%%%%%%%%%%%%
However, the improved efficiency achieved through kernelization often comes with additional, sometimes impractical assumptions on the attention matrix \cite{wang2020linformer} or with approximations of the softmax operation under specific theoretical conditions \cite{choromanski2023learning}. However, when these assumptions are not satisfied or approximation errors occur, the performance of these methods may not be consistently better than vanilla transformer. For example, transformer variants with linear complexity, such as ScatterBrain \cite{chen2021scatterbrain}, and KDEformer \cite{zandieh2023kdeformer}, show less satisfactory performance on ImageNet-1k compared to Swin \cite{liu2021swin}, as observed in Figure \ref{kernel}. 
%%%%%%
%%%
%%%
Here, we emphasize that while softmax is computationally expensive, but it is an accurate function for calculating attention weights. Given this challenge, a fundamental question arises: ``{\it Is there an alternative to DPSA that maintains softmax's properties while efficiently modeling long-range interactions}?".  
%%%
In softmax attention, two factors play a vital role in its performance: non-negative elements within the attention matrix \cite{zandieh2023kdeformer}, and a nonlinear reweighting mechanism that enhances the stability of attention weights \cite{gao2017properties, zhu2021long}. 
%%%%%%%%%%%%%%%%%%%%%%%%%%%%%%%%%%%%%%%%%%%%%
We propose a new variant of DPSA called Self-optimal Transport (SeT) that satisfies both of these properties by using principles from optimal transport (OT) and kernel method. In SeT, input features are mapped into a Reproducing Kernel Hilbert Space (RKHS), in which point evaluation takes the form of a linear function \cite{bietti2019group}. Instead of using dot-product to find the similarity score between elements, we adopt alignment scores achieved through the computation of optimal transport between input and reference features, using Sinkhorn's method \cite{cuturi2013sinkhorn}.
%%%
%%
%we use the alignment scores that are efficiently derived by computing the transport plan between the input and the reference sets through Sinkhorn's algorithm \cite{cuturi2013sinkhorn}. The references serve as anchor points to guide the alignment/aggregation process. 
%%%
%In contrast to utilizing dot-products for element similarity, we adopt alignment scores acquired through Sinkhorn's algorithm \cite{cuturi2013sinkhorn} to efficiently compute the transport plan between input and reference sets.
%%
%%%
%More precisely, we use the kernel method to define a similarity function between tokens, where the OT allows us to efficiently compute the alignment between them. %In the other words, {\it the sophisticated attention matrix can be obtained as a by-product of KOT}. 
%
%Importantly, unlike some of the kernel-based approaches \cite{choromanski2020rethinking}, our method can work with positional encoding techniques as well. 
%%%%
OT, is an effective computational technique for aligning distributions, which  has found significant attention in many applications, including computer vision \cite{de2023unbalanced}, and machine learning \cite{liu2022sparsity}. As the name indicates, OT aims to find the most efficient way for transporting a mass from one location to another, while in our model, it efficiently computes similarity weights by aligning each feature in the input with a reference set. %By using SeT, we can better capture the relationships and dependencies among input, leading to more efficient computations. 

\noindent{\bf{Contributions.}} SeT involves a two-step process: constructing the kernel feature map and then applying OT to compute the alignment matrix. We begin by embedding input feature vectors into a RKHS, where a positive definite kernel ensures the non-negative property. This enables the model to avoid aggregating negatively-correlated information. The non-linear reweighting scheme is achieved by assigning different weights to tokens based on their relevance with the reference set. OT aligns input feature vectors with a reference set, enhancing local correlations and capturing complex dependencies.  To maintain computational feasibility, we use a kernel approach \cite{williams2000using} to obtain a finite-dimensional embedding, and multi-level hierarchical representations, similar to Swin \cite{liu2021swin}, are constructed through kernel compositions, resulting in downsampled feature maps across levels.
%%%%%
%%
Our new transformer, SeTformer, demonstrates its effectiveness through extensive experiments on various tasks. This highlights the potential of content-based interactions to enhance transformer performance in vision and language applications.

\section{Related Works}
%Given the vast literature on improving the efficiency of transformers, and particularly the DPSA bottleneck, we focus on the most related recent works.  
%%%%
{\bf{Vision Transformers.}} Transformer and self-attention mechanism have significantly impacted  Natural Language Processing \cite{vaswani2017attention}, and their application to vision tasks has been made possible by the pioneering Vision Transformer (ViT) \cite{dosovitskiy2020image}. Researchers have further  extended ViT models in multiple directions, focusing on position encoding \cite{chu2021conditional}, data efficiency \cite{touvron2021training}, and optimization techniques \cite{li2112improved}. These advancements have led to significant progress in vision tasks \cite{khan2022transformers}. Several recent works have focused on enhancing ViT' performance on downstream tasks by exploring pyramid structures, surpassing convolution-based methods. PVT \cite{wang2021pyramid, wang2022pvt} introduces sparse location sampling in the feature map to form key and value pairs. Swin \cite{liu2021swin} utilizes non-overlapping windows with window shifts between consecutive blocks. CSwin \cite{dong2022cswin}  extends this approach with cross-shape windows to enhance model capacity. PaCa-ViT \cite{grainger2023paca} introduces a new approach where queries start with patches, while keys and values are based on clustering, learned end-to-end. HaloNet \cite{vaswani2021scaling} introduced a haloing mechanism that localizes self-attention for blocks of pixels instead of pixel-wise, aiming to address the lack of an efficient sliding window attention. Similarly, NAT \cite{hassani2023neighborhood} adopts neighborhood attention, considering specific scenarios for corner pixels. FocalNets \cite{yang2022focal} replaces self-attention with a Focal module using gated aggregation technique for token interaction modeling in vision tasks. While, these advances have been successful and achieved impressive results on vision tasks, they often come with higher complexity compared to vanilla ViT counterparts. With SeTformer, we've aimed to achieve a balance between performance and complexity, addressing this issue.

\noindent{\bf{Kernel Methods for Transformers.}} When dealing with large-scale input, a more efficient approach is to directly reduce the complexity of the theoretical calculations. Kernelization accelerates self-attention by transforming the computation complexity from quadratic to linear. By utilizing kernel feature maps, we can bypass the computation of the full attention matrix, which is a major bottleneck in softmax. %This provides a path for developing more efficient attention mechanisms.
%%%%%%%%%%%%%%%%%%
%Kernelization offers another efficient principle. When faced with longer input sizes, using kernel-based attention avoids the explicit computation of the attention matrix, which is the bottleneck for softmax transformers \cite{chowdhury2022learning}. 
%%%%%%%%%%%%%%%%%%%%%%%%%%%%%%%%%%%%%%%%%%%%%%%%%%%%%%%%
Recent advances in scalable transformers include  \cite{choromanski2020rethinking, peng2021random, chowdhury2022learning, choromanski2023learning, zandieh2023kdeformer}, where the self-attention matrix is approximated as a low-rank matrix for long sequences. These methods are either very simplistic \cite{chowdhury2022learning}, while others are mathematically well explained but complex  \cite{choromanski2020rethinking, zandieh2023kdeformer}. Several of these methods rely on the Fourier transform, leading to sin and cos random features, which are unsuitable for transformers due to negative values in the attention matrix. To address this, \citet{choromanski2020rethinking} proposed a solution using positive valued random features known as FAVOR+ for self-attention approximation. This approach was improved by \citet{likhosherstov2022chefs}, through carefully selecting linear combination parameters, allowing for one parameter set to be used across all approximated values. While all these methods achieved decent efficiency, they often falls behind popular vision transformers like Swin in terms of performance (Figure \ref{kernel}). 
%%%%%%%%%%%%%%%%%%%%%%%%%%
% Another challenge with kernel-based approaches is their incompatibility with positional encoding (PE) techniques. PE typically requires the explicit computation of the attention matrix, which is exactly what these kernel-based methods avoid to improve the efficiency. Despite some individual works \cite{hedegaard2022continual, chowdhury2022learning,  qin2023linearized}, general principles for incorporating PE into the kernel transformers are not well-studied. 
%%%%%%%%%%%%%%%%%%%%%%%%
%%%%%%%%%%%%%%%%%%
% \cite{choromanski2023learning} and \cite{hedegaard2022continual} combine implicit attention with PEs for language modeling. In our experiments (Sec. ) some of these methods perform well for sequence modeling, but faced memory issues when applied to image classification.
%%%%%%%%%%%%%%%%%%%%%%%%%%%%%%%%%%%
%%%%%%%%%%%%%%%%
%One of the reasons for this performance limitations is that the similarity functions/kernels, like the scaled-dot-product, are typically hand-picked and not learned from data. %Thus, striking a balance between accuracy and computational requirements remains a significant challenge an optimal transformer. 
However, we explore the concept of optimal transport and kernel learning to potentially overcome this performance gap while still maintaining the efficiency advantages of Transformers. 

%%%%%%%%%%%%%%%%%%%%%%%%%%%%%%%%%%%%%%%%%%
Our model's corresponding kernel is a matching kernel \cite{tolias2013aggregate}, which uses a similarity function to compare pairs of features (the input vector and reference set). Recent methods have also explored kernels based on matching features using wasserstein distance \cite{khamis2023earth, kolouri2020wasserstein}. Prior studies by \citet{skianis2020rep} and \citet{mialon2020trainable} analyze similarity costs between input and reference features in biological data. They employ dot-product operation \cite{vaswani2017attention} for element-wise comparison. In contrast, our model employs transport plans to compute attention weights, providing a new perspective on Transformer’s attention via kernel methods. %Additionally, a similar construction is used in the recent study \cite{clark2022unified}, where the attention matrix is replaced in \cite{lewis2021base} by the effective optimal transport to approximate matching steps during expert selection. 
Unlike other kernel-based methods like Performer that is not compatible with positional encoding techniques, our model incorporates this information, which is crucial in visual modeling tasks.  
\section{Proposed Method} \label{sec:3}
\subsection{From Self-Attention to Self-optimal Transport}
Consider an input sequence $x\text = \{x_1,...,x_n\}$ $\in$ $ R^d$ of $n$ tokens. The DPSA is a mapping that inputs matrices $Q, K, V \in R^{n\times d}$. These matrices are interpreted as queries, keys, and values, respectively,
\begin{equation}
 \begin{aligned}
Att (Q, K, V) &= D^{-1} A V \\  A = \exp \Big( {Q K^T}/{\sqrt{d}} \Big) &;\;\; \; D = \mbox{diag}(A{\bf{1}}_n^T)
%Att (Q, K, V) = softmax \Big[ \frac{Q K^T}{\sqrt{d}} \Big] V
\label{eq1}
\end{aligned}
\end{equation}
where $\bf{1_n}$ is the ones vector, and $A, D \in R^{n\times n}$. The Softmax operation normalizes the attention weights {\bf A}, allowing each token to be a weighted average of all token values. However, the quadratic complexity of the softmax becomes a bottleneck as the number of tokens increase. 
%%%%%%%%%%%%%%%%%%%%
%As demonstrated by [18], two essential properties of the softmax operation play crucial roles in its performance: (i) ensures all values in the attention matrix are non-negative; (ii) provides a nonlinear reweighting mechanism, concentrating the distribution of attention connections and stabilizing the training process.
%%%%%%%%%%%%%%%%%%%%
%%%%%%%%%%%%%%%%%%%%%%%%%%
%The principle of DPSA relies on the pairwise similarity between inputs in the vector space -- i.e., high similarities assign strong attendance to the corresponding input tokens. Time and space complexity of computing (\ref{eq1}) are $\mathcal{O}(n^2d)$ and $\mathcal{O}(n^2+nd)$ respectively, because {\bf{A}} has to be stored explicitly. 
%%%%%%%%%%%%%%%%%%%%%%%%%%%%%%%%%%%%%%%%
We aim to develop a powerful and efficient self-attention that is, above all, simple. We do not add any complex modules like convolution \cite{wu2021cvt}, shifted windows \cite{hassani2023neighborhood}, or attention bias \cite{li2022next} to improve the vision task's performance. We indeed take a different strategy. SeT leverages the important properties of softmax, including non-negativity and reweighting mechanism \cite{gao2017properties}, while also prioritizing efficiency in its design. The use of RKHS with a positive definite (PD) kernel avoids aggregating negative-correlated information. SeT incorporates a nonlinear reweighting scheme through OT. This involves computing alignment scores between input and reference sets within RKHS. This process introduces nonlinearity to the alignment scores, assigning weights to elements to highlight their significance. This helps the model in capturing complex relationships and emphasizing local correlations. %Additionally, OT inherently enforces non-negativity in the alignment process by optimizing the transport plan between input features and a reference set \cite{sinkhorn1967concerning, cuturi2013sinkhorn}. This ensures that attention weights remain non-negative, aligning with the properties of softmax attention.

%Additionally, the OT also enforces non-negativity in the alignment process, by finding the most efficient transport plan between input features and a reference set \cite{sinkhorn1967concerning, cuturi2013sinkhorn}. The references are the anchor points, guiding the alignment process. These process are detailed in below and Figure \ref{fig2}. 

%In this paper, we show that by using our methodology, we can reduce the overall complexity from $\mathcal{O}(n^2)$ to $\mathcal{O}(n m)$, where $m$ is the number of references. An important benefit of using a reference set is to reduce the size of  the attention matrix, which improves the practical performance on large datasets. 
%
\begin{figure}[t]
\centering
 \includegraphics[width=8.5cm]{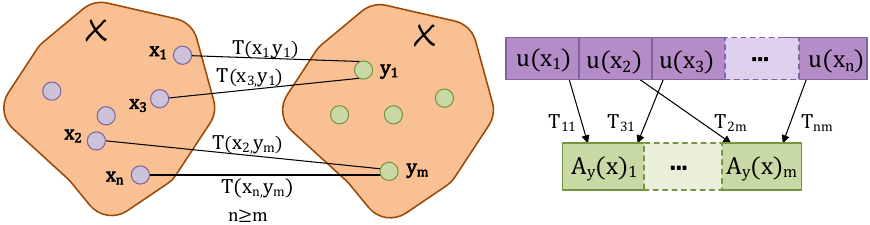}
 %\vspace{-6pt}
 \caption{An input feature vector $x$ is transported onto reference $y$ via  transport plan $T(x, y)$, that aggregates $x$ features w.r.t. $y$, yielding $A_y(x)$. In DPSA, each $x_i$ aggregates with all $x$ features, forming a large sparse matrix. Our model aggregates based on best-matched $x$ and $y$ features through OT.}
\label{fig2}
\end{figure}

%\subsection{Preliminaries}
% We introduce a way to learn self-attention, which combines principles from  \cite{peyre2019computational} with kernel methods \cite{scholkopf2002learning}. Given an input image transformed into a set of feature vectors $\{\mathcal{X}= x|\{x_1,...,x_n\}$ such that $x_1,...,x_n \in R^{d}$ for $n\geq1\}$. Our model, KOT, builds upon two main ideas that are detailed below and also illustrated in Figure \ref{fig2}. \\
%\vspace{-10pt}

\subsubsection{Representing local image neighborhoods in an RKHS.} In order to maintain a linear computation, we embed the input feature vector into a RKHS, in which point evaluation takes the form of linear function \cite{jagarlapudi2020statistical}. %RKHS allows us to construct a powerful and flexible functional space for learning and representing intricate patterns and dependencies that regular attention mechanisms may struggle to capture \cite{bietti2019group}. 
Kernel methods enable us to map data from its original space $\mathcal{X}$ to a higher-dimensional Hilbert space (feature space) $\mathcal{F}$ through a positive definite (PD) kernel $\mathcal{K}$ \cite{zhang2019optimal}. For a function $u$: $\mathcal{X}\rightarrow \mathcal{F}$ (feature map), the PD kernel is denoted as $\mathcal{K}(x, x')=\langle u(x), u(x')\rangle_\mathcal{F}$. Given that $u(x)$ can be an infinite-dimensional, the kernel technique \cite{williams2000using} allows to derive a finite-dimensional  representation $v(x)$ in $R^k$, with an inner product $\langle v(x_i), v(x'_j)\rangle$ denoting $\mathcal{K}(x, x')$. As shown by \citep{fukumizu2008elements}, if $\mathcal{K}$ is positive definite, for any $x$, $x'$, we have $\mathcal{K}(x, x')\ge0$, which aligns with the non-negativity property of softmax operator.

%More precisely, the input elements are aggregated through a weighted sum, where the weights are computed using optimal transport.

\subsubsection{Optimal transport (OT).} A fundamental role in our model is to aggregate related tokens by learning a mapping between them. Our weighted aggregation relies on the transport plan between elements $x$ and $x'$ treated as distinct measures or weighted point clouds. OT has found extensive use in alignment problems \cite{khamis2023earth}, and has an impressive capacity to capture the geometry of the data \cite{liu2022sparsity}. We focus throughout this paper on the Kantorovich form of OT \cite{peyre2019computational} with entropic regularization for smoothing transportation plans. 
%%%%%%%%%%%%%%%%%%%%%%%%%%%%%%%%%%%%%%%%%%
%We consider that the elements in the input are aggregated together based on their alignment with the reference set. The alignment scores are efficiently obtained by computing the transport plan between the input and reference sets using Sinkhorn's algorithm \cite{cuturi2013sinkhorn}. More precisely, OT \cite{cuturi2013sinkhorn} computes the transport plan between $x$ and $x'$ represented as weighted point clouds or discrete measures.  
%%%%%%%%%%%%%%%%%%%%%%%%%%%%%%%%%%%%%%%%%%%%%%%%
%Instead of using the dot product similarity used in self-attention, OT computes alignment scores by finding the transport plan that minimizes the cost of aligning elements from the input sequence with elements from a reference set. %self-attention uses dot product similarity, while optimal transport computes alignment scores based on transport costs.
%In our model, the input feature vectors will be aggregated if they align well with the corresponding reference. The degree of alignment can effectively be computed via OT. 
%%%%%%%%%%%%%%%%
Let $g$ in $\mu_n$ and $h$ in $\mu_{n'}$ denote the weights of discrete measures $\sum_i g_i \delta_{x_i}$ and $\sum_j h_j \delta_{x'_j}$ for the elements $x$ and $x'$. Here, $\delta_x$ is the  unit mass with $x$. The cost matrix $C\in R^{n\times n'}$ has entries $c(x_i,x'_j)$ for the $(i,j)$ pairs. Instead of computing a pairwise dot product between distributions, OT finds the minimal effort based on the ground cost to shift the mass from one distribution to another, and can be defined by
%\vspace{-10pt}
\begin{equation}
%\begin{aligned}
 OT_{\text{kant}}=\min_{T \in U(g,h)}\sum_{ij} C_{ij}T_{ij} + \epsilon H(T) 
\label{eq:6}
%\end{aligned}
\end{equation}
%\vspace{-15pt}

%Let $x = {x_1,\dots,x_n}$ denote the set of n samples from $\Delta_n$ and let $y = {y_1,\dots,y_m}$ denote m samples from $\Delta_m$. Let a in $\Delta_n$ (probability simplex) and b in $\Delta_m$ be the weights of the discrete measures $\sum_i \delta_{x_i}$ and $\sum_j \delta_{y_j}$ with respective locations x and y, where $\delta_x$ denotes the Dirac (unit mass) at position $x$. Let $C\in R^{n\times m}$ denote a function that evaluates the cost between elements $x$ and $y$. 

\noindent where the negative entropy function is defined as ${H (T)}$ = $\sum_{i,j} T_{ij} (\log (T_{ij}) -1)$ with regularization parameter $\epsilon$. The transport plan $T_{ij}$ describes the amount of mass flowing from location $i$ to location $j$ with minimal cost, and the constraint $U(g, h)\text = \{T\in R_{+}^{n\times n'}: T1_n\text = g; ~ T^T 1_{n'}\text = h\}$ represents the uniform transport polytope. The computation of OT in (\ref{eq:6}) is efficiently done using a matrix scaling procedure derived from Sinkhorn's algorithm \cite{cuturi2013sinkhorn}. OT assigns different weights to individual elements/tokens based on their significance within the input, similar to the reweighting scheme in softmax attention. Furthermore, OT enforces non-negativity by optimizing alignments between input elements \cite{sinkhorn1967concerning, cuturi2013sinkhorn}, preserving the non-negative nature of attention weights, as in softmax attention. %Additionally, OT inherently enforces non-negativity in the alignment process by optimizing the transport plan between pairs of elements within the input \cite{sinkhorn1967concerning, cuturi2013sinkhorn}. This ensures that attention weights remain non-negative, as in softmax attention. 
Indeed, kernels capture the nonlinear transformation of the input, while OT finds optimal alignments between sets of features with fast computation. 

%Our proposed approach leverages the optimal transport and kernel methods to satisfy both the softmax properties  the transport plan between x and x', offering a new perspective on learning weighted aggregation.

%

%To this end, one can use a function $u$: $\mathcal{X}\rightarrow \mathcal{F}$ to map the data living in the space $\mathcal{X}\in R^d$, to a RKHS $\mathcal{F}$ with positive definite kernel $\mathcal{K}$, where,  $\mathcal{K} (x, x')$ = $\langleu (x), u (x')\rangle_\mathcal{F}, ~ \forall x, x'\in \mathcal{X}$. However, the data $u(x)$ is approximated by vectors that may have varying sizes, thus, this issue can be solved by using kernel approximation method \cite{williams2000using} which allows extracting a fixed-size representation $v(x)$ in $R^k$, and $\langle v(x_i), v(x'_j)\rangle$:= $\mathcal{K}(x, x')$. 

% In fact, the input features aggregated if match/align well with the corresponding reference through the transport plan. With such a combination, we propose an efficient variant of SA that can operate on features of varying sizes, and is compatible with a broad range of vision tasks. 
%and works better or similar to the high-performing convolutional models while having less computation complexity. 
%The elements of $x$ can be a vector representations of image patches, or words for sentences.

\vspace{-3pt}

\subsection{Self-optimal Transport (SeT)}
\begin{figure}
\centering
 \includegraphics[width=8.2cm]{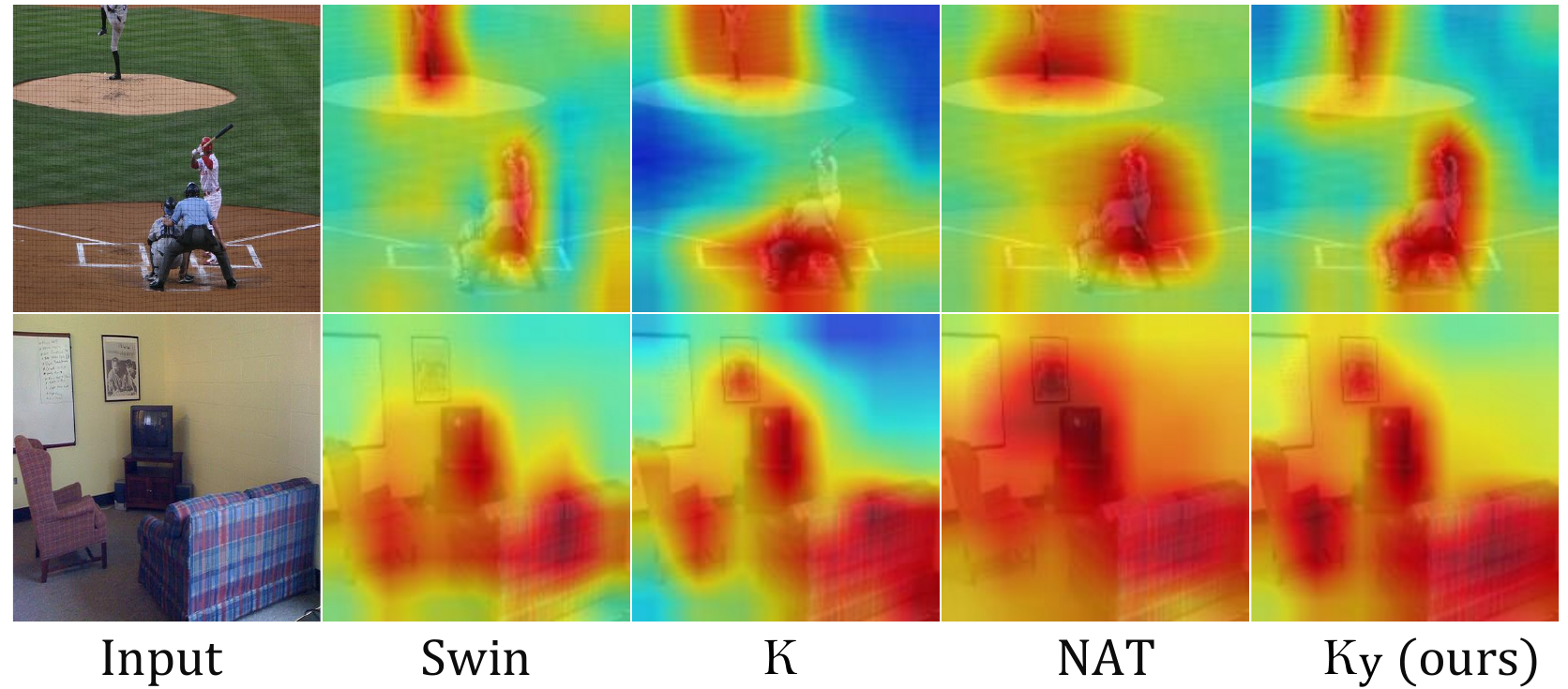}
 \caption{Visualization of our model on COCO dataset using $\mathcal{K}$, and $\mathcal{K}_y$. They are compared with the Swin \cite{liu2021swin} and NAT \cite{hassani2023neighborhood}. While $\mathcal{K}$ focuses solely on large objects, $\mathcal{K}_y$ accurately captures multi-scale objects without including sparse areas.}
\label{fig4}
\end{figure}

%We propose SeTformer, which preserves the non-negativity and re-weighting characteristics of the softmax while completely discarding the DPSA. 
With an input feature vector $x$ and a reference $y_m$ in $\mathcal{X}$, we perform the following steps: (i) representing the feature vectors $x$ and $y$ in $\text{RKHS}_{\mathcal{F}}$, (ii) aligning the elements of $x$ with $y$ using OT, (iii) performing a weighted aggregation of the elements $x$ into $m$ clusters, resulting in an alignment matrix $A$, as detailed in Figure \ref{fig2}. 
%%%%%%%%%%%%%%%%%%%%%%%%%%%%%%%%%%%%%%%%%%%%%%%%%
We use a reference $y$ for efficient element aggregation. Each element in the reference set serves as an ``alignment cell",  and input features are aggregated within these cells through a weighted sum.  These weights indicate the correspondence between the input and references, computed using OT. 
%%%%%%%%%%%%%%%%%%%%%%%%
Suppose we have an input feature vectors $x$ = $\{x_1,\dots, x_n\}$ living in $\mathcal{X}\in R^d$, randomly extracted from input images. In the context of the Nystr\"om approximation \cite{xiong2021nystromformer}, the samples of $y$ is the centroids obtained by conducting K-means clustering to feature vectors in the training set $\mathcal{X}$, such that we obtain $y$ = $\{y_1,\dots, y_m\}$ with $m\leq n$. The use of the reference set helps optimize the computation process and enables the model to scale effectively for longer input sequences \cite{skianis2020rep, mialon2020trainable}.
%%%%%%%%%%%%%%%%
Let $k$ be a positive-definite kernel, like the Gaussian kernel that defined on RKHS along with mapping $u: R^{d}\rightarrow \mathcal{F}$. We create a matrix $k$ of size $n\times m$ that stores the comparisons $k(x_i, y_j)$. Next, we compute the transport plan between $x$ and $y$ based on (\ref{eq:6}), resulting in the $n\times m$ matrix $T(x, y)$. The transport plan finds the best way to align the input features with the reference elements while minimizing the alignment cost. Our $A_y(x)$ is now defined as
%%%%%%%%%%%%%%%%%%%%%%%%%%%%%%%%%%%%%%%%%
\begin{equation}
\begin{aligned}
m^{1/2}  \Big[\sum_{i=1}^n T(x, y)_{i 1} u (x_i),\dots,\sum_{i=1}^n T(x, y)_{i m} u (x_i) \Big]= \\ 
m^{1/2}~ T(x, y)^T u (x) \text{ where } u(x)=[u(x_1),\dots,u(x_n)]^T.
\label{eq:emb}
\end{aligned}
\end{equation}
%%%%%%%%%%%%%%%%%%%%%%%%%%%%%%%%%%%%%%%%%%%%%%%%%%%%%%%%%%%%%%%%%%%%%%%%%%%%%%%%
This aggregation relies on a positive-definite kernel such that   
\vspace{-5pt}
\begin{equation}
\mathcal{K}_y(x, x')=\sum\limits_{i, i'=1}^n  T_y(x, x')_{i i'} k (x_i, x'_{i'})
\label{eq:4}
\end{equation}
%\vspace{-2pt}
%%%%%%%%%%%%%%%%%%%%%%
This formulation with $T_y(x, x')$ = $m\cdot T (x, y) T(x', y)^T$, can be seen as a kernel that assigns weights to match between pairs of elements. The weights are determined by OT in the feature space ($\mathcal{F}$). Our attention process involves aggregating the non-linearly embedded elements of $x$ into ``cells", with each cell corresponding to a reference element in $y$, guided by the transport plan $T(x, y)$. The reweighting property in $A_y(x)$ arises from using OT to align the non-linearly embedded input $x$ with the reference $y$. The weights assigned during aggregation, driven by this alignment, are guided by the OT process. In other words, each value in $T_y(x, x')$ represents the weight or importance of $x$ aligned with the elements of $x'$. Since the OT matrix is constructed from non-negative values \cite{cuturi2013sinkhorn, jagarlapudi2020statistical}, the aggregation process will involve non-negative weights, satisfying the non-negativity property. 
%%%%%%%%%%%%%%%%%%%%%%%%%%
However, if we exclude the use of a reference for cost calculation, we will have $\mathcal{K}(x, x')$ = $\sum_{i, i'=1}^n T(x, x')_{i i'} k(x_i, x'_{i'})$, which is computationally costly due to the quadratic number of transport plans required.  To differentiate between this kernel and $\mathcal{K}_y$, we need to find the relationship between $T_y(x, x')$ and $T(x, x')$, which is elaborated in the {\it Suppl. file}. Consider $x, x', y$ with lengths $n, n'$, and $m$ respectively. The weighted wasserstein distance $\mathcal{W}_y(x, x')$ based on reference $y$ is $\langle T_y (x, x'), d_k^2(x, x')\rangle^{1/2}$, where $d_k^2(x, x')$ is a distance metric induced by the kernel. 
%%
%%
%%%
The inequality $|\mathcal{W}(x, x') - \mathcal{W}_y (x, x')| \leq 2min (\mathcal{W} (x, y), \mathcal{W}  (x', y))$, shows that the weighted wasserstein distance $\mathcal{W}_y(x, x')$ is a valid approximation of the actual wasserstein distance $\mathcal{W}(x, x')$. The difference between these two distances is bounded by a factor of $2$ times the minimum of the wasserstein distances between the input sets $x$ and $y$; $x'$ and $y$. %Indeed, the weighted wasserstein distance allows for faster computation compared to the actual wasserstein distance.
%%%%%%%%%%%%%%%%%%%%%%%%%%%%%%%%%%%%%%%
%On the other hand, $T_y(x, x')$ is a rough approximation of the optimal coupling $T (x, x')$. Accordingly, $\mathcal{K}_y$ can be a solution to the transportation problem and defines as $\mathcal{K}_{T} (x, x')$ = $\sum\limits_{i, i'=1}^n T(x, x')_{i i'} k(x_i, x'_{i'})$. $\mathcal{K}_T$ on the other hand is equal to the Wasserstein distance ($\mathcal{W}$ for short), with distance metric $d_k$ induced by $k$. Pey\'er et al. \cite{peyre2019computational} show that $\mathcal{K}_{T}$ relies on kernels that are not admitting to positive-definiteness and are computationally expensive (the proof of which can be found in \cite{peyre2019computational}, chapter 8.3), while the distance with the $\mathcal{K}_y$ is a simple iterative scheme with linear convergence. 
%%%%%%%%%%%%%%
This efficient computation is highlighted in Figure~\ref{fig4} where $\mathcal{K}_y$ effectively captures objects of different scales, emphasizing essential features of each object rather than the surrounding region. Indeed, $\mathcal{K}$ mainly focuses on larger objects (e.g., sofa), and $\mathcal{K}_y$ accurately captures smaller objects like cricket sticks. NAT \cite{hassani2023neighborhood} excels in capturing multi-scale objects but also include the sparse region around each object.  
%%%%%%%%%
\begin{figure*}
\centering
 \includegraphics[width=17.7cm]{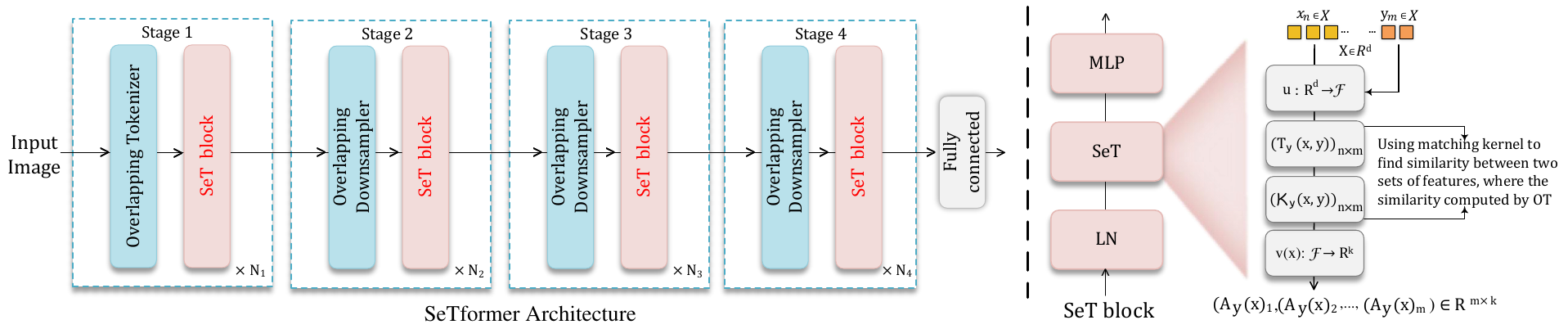}
 \caption{SeTformer Architecture (Left). It starts with a down-sampling convolutional layer, followed by four sequence stages containing multiple SeT blocks. Consecutive stages are bridged by down-sampler layers that reduce the spatial size while doubling the depth. On the right, we illustrate our attention formulation: mapping $x$ and $y$ elements to the RKHS, then aggregating $x$ features if they align well with the corresponding reference through OT computation between $x$ and $y$.}
\label{fig3}
\end{figure*}

%%%%
\subsubsection{Projecting onto a linear subspace}
When dealing with finite-dimensional $u(x)$, the $A_y(x)$ can be directly computed without incurring significant computational overhead. In the case of infinite or high-dimensional $u(x)$, the Nystr\"{o}m algorithm \cite{williams2000using, xiong2021nystromformer} provides an efficient approximation for the embedding $v: {R}^d\rightarrow{R}^k$. The  Nystr\"{o}m algorithm  approximates the transport plan by sampling a subset of columns and rows, and projecting input from the feature space $\mathcal{F}$ onto a linear subspace $\mathcal{F}_1$, which results in an embedding $\langle v(x_i), v(x'j)\rangle{\mathcal{F}_1}$. The subspace $\mathcal{F}_1$ is spanned by $k$ centroids $u(z_1), \ldots, u(z_k)$.
%%%%%
The explicit formula $v(x_i) = k(z, z)^{-1/2}  k(z, x_i)$ represents a new embedding with $z = {z_1, \ldots, z_k}$ as centroids. This efficient method only requires performing K-means clustering and computing the inverse square root matrix. To implement this approximation, replace each $x_i$ with its corresponding embedding $v(x_i)$, updating the attention model in (\ref{eq:emb}),
%%
%%%%%%
%%%%%%%%%%%%%%%%%%%%%%%%%%%%%%%%%
\vspace{-3pt}
\begin{equation}
\begin{aligned}
m^{1/2} \Big[\sum_{i=1}^n T(v(x), y)_{i 1} v(x_i),\dots,\sum_{i=1}^n T(v(x), y)_{i m} v(x_i)\Big] \\
=m^{1/2}~T(v(x), y)^T v(x). 
\end{aligned}
\label{eq:appp}
\end{equation}
%%%%%%%%%%%%%%%%%%%%%%%%%%%%%%%%%%%%%%%%%%%%%%%%%%
here, $v(x)\in R^{m\times k}$ with $m$ denoting the features in $y$. 
\subsubsection{Linear positional encoding}
\label{pos}
To incorporate the positional information into our model, we follow \cite{mairal2016end} and apply an exponential penalty to the similarities between the input and reference sets based on their positional distance.
This involves performing a multiplication of $T(v(x), y)$ with a distance matrix $\mathcal{M}$, where  $\mathcal{M}_{i j}=e^{(-\frac{1}{\tau^2})(\alpha-\beta)^2}$, with $\alpha=i/n$, $\beta=j/m$, and $\tau$ representing the smoothing parameter. Subsequently, the updated value is replaced into (\ref{eq:appp}). 
Our similarity weights that consider both content and positional information achieves superior performance compared to other positional encoding methods (see Table \ref{tab:ab}). 

\begin{table}[h]
\centering
\small{
\begin{tabular}{llcc} 
{Model} & {Layers} &{\# Param.} & {FLOPs}    \\   \hline 
SeTformer-Miny & 3, 4, 6, 5 & 16M & 2.8G  \\ [-0.4ex]
SeTformer-Tiny & 3, 4, 18, 5  & 22M & 4.1G \\  [-0.4ex]   
SeTformer-Small & 3, 4, 18, 5 & 35M & 6.9G \\    [-0.4ex] 
SeTformer-Base & 3, 4, 18, 5  & 57M & 10.2G 
\end{tabular}}
\vspace{-5pt}
\caption{Different variants of the SeTformer architecture.} \label{conf}
\end{table}

\definecolor{Gray}{gray}{0.8}

\setlength{\tabcolsep}{2.5pt}
\begin{table}[t]
\centering
\small{
\begin{tabular}[t]{@{}l|cccl@{}}   
{Model} & {\# Param.} & {FL.} & {Thro.} & {Top-1}  \\   \hline  
%& (M) & (G)& (image/s) & (\%)\\ \hline  \hline 
%ResNet-50 &  25.0M   & 4.1G  & 1257/s &  76.8 \\[-0.4ex]
%ResNet-101 &  46.0  & 7.9  & 746 & 77.5\\[-0.2ex] \hline
\rowcolor{Gray}
\bf SeTformer-M & 16.2M  & 2.8G  & 793/s & 81.7 \\ \hline
%PaCa-T & 12.2M & 3.2G & - & 80.9 \\ 
Swin-T \cite{liu2021swin} &  28.2M  & 4.5G  & 755/s & 81.3    \\   [-0.3ex]
%PVT-S &  24.9M & 3.8G & 820/s & 79.7 \\ [-0.3ex]
ConvNeXt-T \cite{liu2022convnet}& 28.4M  & 4.5G   & 775/s & 82.1\\  [-0.3ex]
%\textcolor{cyan}{$\circ$} Shunted-T \cite{} & 11.5 &  2.1 &  736 &  79.8 \\  [-0.3ex]
FocalNet-T \cite{yang2022focal}& 28.6M  & 4.5G   &696/s & 82.3 \\  [-0.3ex]
CSWin-T \cite{dong2022cswin}& 22.8M & 4.3G & 701/s & 82.7 \\ [-0.3ex]
%PaCa-S &  21.1M &  5.4G & - & 83.17 \\ [-0.3ex]
MViTv2-T \cite{li2112improved} & 24.0M & 4.7G & 776/s & 82.3 \\  [-0.3ex] 
NAT-T \cite{hassani2023neighborhood} & 28.0M & 4.3G & 752/s & \underline{83.2} \\ [-0.3ex] \rowcolor{Gray}
\bf SeTformer-T & 22.3M  & 4.1G  &  785/s & \bf83.9 {\tiny{(+0.7)}}  \\ \hline
Swin-S \cite{liu2021swin} & 49.6M  & 8.7G  & 437/s & 83.0\\  [-0.3ex]
ConvNeXt-S \cite{liu2022convnet} & 49.5M  & 8.7G  & 448/s & 83.1\\  [-0.3ex]
FocalNet-S \cite{yang2022focal} & 50.3M  & 8.7G  & 406/s & 83.5\\  [-0.3ex]
CSWin-S \cite{dong2022cswin}& 35.1M &  6.9G &  437/s &  83.6 \\ [-0.3ex]
MViTv2-S \cite{li2112improved}& 35.0M & 7.0G & 341/s & 83.6 \\ [-0.3ex] 
NAT-S \cite{hassani2023neighborhood} & 51.0M & 7.8G & 435/s & \underline{83.7} \\  [-0.3ex]\rowcolor{Gray}
\bf SeTformer-S  & 35.0M & 6.9G & 451/s & \bf 84.7 {\tiny{(+1.0)}}\\ \hline

%DeiT-B  & 86.3M  & 17.5G  & 292/s & 81.8\\  [-0.3ex]
%PVT-L  & 61.3M & 9.8G & 367/s & 81.7 \\ [-0.3ex]
Swin-B \cite{liu2021swin} & 87.4M & 15.4G &  278/s &  83.3   \\    [-0.3ex]
ConvNeXt-B \cite{liu2022convnet} & 88.9M & 15.4G  & 297/s & 83.5    \\    [-0.3ex]
FocalNet-B \cite{yang2022focal}  & 88.7M  & 15.4G  & 296/s  & 83.9    \\   [-0.3ex]
CSWin-B \cite{dong2022cswin} &  78.2M & 15.2G &  250/s  &  84.2 \\  [-0.3ex]
%PaCa-B & 46.7M & 9.7G & - & 84.22 \\  [-0.3ex]
MViTv2-B \cite{li2112improved} & 52.0M & 10.2G & 253/s & {84.4} \\ [-0.3ex] 
NAT-B \cite{hassani2023neighborhood}& 90.0M & 13.7G & 294/s & \underline{84.4} \\  [-0.3ex] \rowcolor{Gray}
\bf SeTformer-B  & 56.7M & 10.2G & 298/s  & \bf86.2  {\tiny{(+1.8)}}
\end{tabular}}
%\vspace{-2pt}
\caption{Classification accuracy on the ImageNet-1K at 224$\times$224 resolution.}%Throughput is measured using Pytorch image method \cite{wightman10pytorch} on a V100 GPU.} 
\label{tab:clas}
\end{table}
%%%%%%%%%%%%%%%%%
\setlength{\tabcolsep}{1pt}
\begin{table}
%\vspace{-3mm}
\centering
\small{
\begin{tabular}{@{}l|cc@{}}  
Model & Top-1 acc. & GFLOPs  \\  [-0.3ex]  \hline 
%{$\circ$} ResNet-101  & 86.2M & 1029G & 44.9 / - \\ \hline
   %\rowcolor{gray}
Baseline \cite{yuan2021tokens} & 82.55 &161.10 \\  \hline 
Performer \cite{dosovitskiy2020image}&80.50 & 5.06 (31.87$\times$) \\ [-0.3ex] 
Reformer \cite{kitaev2020reformer}&81.44 &11.71 (13.75$\times$) \\ [-0.3ex] 
ScatterBrain \cite{chen2021scatterbrain}&81.95 &7.18 (22.43$\times$) \\ [-0.3ex] 
KDEformer \cite{zandieh2023kdeformer} &82.08 &8.80 (18.30$\times$) \\ [-0.3ex] \rowcolor{Gray}
\bf SeTformer & \bf 83.17 & 6.9 (\bf23.35$\times$) 
\end{tabular}}
\vspace{-2pt}
\caption{Classification results on the ImageNet-1K adopting T2T-ViT, with the kernel-based methods.} 
\label{kern}  
\end{table}

\subsubsection{SeTformer}
We use Swin as our baseline model, replacing its self-attention with our SeT module. Our model consists of four stages, each with a different spatial resolution, resulting in a spatial size 1/4th of the input image. Inputs are embedded using two-layer 3$\times$3 convolutions with 2$\times$2 strides. After each stage, except the last, there's a down-sampler via 3$\times$3 convolutions with 2$\times$2 strides. This is different from Swin, which uses non-overlapping 2$\times$2 convolutions. The overall network architecture is presented in Figure~\ref{fig3}, and different variants of our model are reported in Table~\ref{conf}.

%%%%%%%%%%%%%%%%%%%%%%%%
\section{Experimental Validation}
We conduct experiments on both image and language domains, including ImageNet, COCO, and ADE20K, as well as the GLUE to demonstrate the impact of our model. We fine-tuned hyper-parameters like the number of references ($m$), entropic regularization $\epsilon$ in OT, and $\tau$ in position embedding. We observed that $\epsilon$ and $\tau$ showed stability across tasks, while, careful selection is required for the value $m$. More details for each dataset are provided in the {\it Suppl. file}. 

%We use the timm \cite{wightman10pytorch} toolkit for classification and mmdetection and mmsegmentation \cite{} for object detection and semantic segmentation, respectively.  

%The ablation studies are performed to characterize the critical elements of our model. 

%\pdfliteral{0.52, 0.52, 0.51 rg}{46.7} \pdfliteral{0 0 0 rg}
\definecolor{Gray}{gray}{0.8}
%>{\color{Gray}}c >{\color{Gray}}c

\setlength{\tabcolsep}{2pt}
\begin{table}[t]
\small{
%{\centering (Throughput also are measured with the same resolution using \cite{}, and a V100 GPU.)}
\begin{tabular}[t]{@{}l| c cc | c cc |c c @{}}  
%\rowcolor{Gray}
\multicolumn{9}{c}{ Mask R-CNN - 3x schedule} \\ \hline 
Model & $\text{AP}^b$ &$\text{AP}_{50}^b$ & $\text{AP}_{75}^b$ &$\text{AP}^m$  &$\text{AP}_{50}^m$  & $\text{AP}_{75}^m$ & Param. & FL. \\ \hline  
%& (M) & (G) & \\ 
%\multicolumn{9}{c}{Mask R-CNN, ~ $3\times$ schedule}  \\ \hline  
ResNet-50   & 41.2 & 61.8 & 44.9 & 37.2 & 58.4 & 40.1 & 44M & 260 \\[-0.3ex]
ResNet-101  & 42.7 & 63.2 &  47.5&  38.4 & 60.3 & 41.5  & 63M & 335 \\ \hline 
\rowcolor{Gray}
\bf SeTformer-M & \bf46.7 & 68.2 & 51.7 & \bf41.9 & 65.1 &  44.7 & 32M & 190   \\  \hline  
%%%%%%%%%%%%%%%%%%%%%%%%%%%%%%%%%%%%%%%%%%%%%%%%%%%%%%%%%%%%%%%%%%
%PVT-S  & 44 & 245 & 43.1 & 65.3 & 46.9 & 39.9 & 62.5 &  42.8 \\[-0.3ex]
Swin-T  & 46.0 & 68.2 &  50.4 & 41.6 & 65.2 &  44.8  & 48M & 264   \\[-0.3ex]
%ConvNeXt-T   & 46.2 & 67.5 & 50.8 & 41.7 & 65.1 & 44.9 & 48M & 253  \\ [-0.3ex]
FocalNet-T  & 48.0 & 69.7 & 53.0 & 42.9 & 66.5 & 46.1  & 49M  & 268    \\[-0.3ex]
CSWin-T & 49.0 & 70.7 & 53.7 & 43.6 & 67.9 &  46.6 & 42M & 279  \\[-0.3ex]
NAT-T &  47.7 & 69.0 & 52.6 & 42.6 & 66.1 &  45.9  & 48M &  258 \\[-0.3ex]
MViTv2-T &  48.2 & 70.9 & 53.3 & 43.8 & 67.9 & 47.2 & 44M & 279  \\[-0.3ex] \rowcolor{Gray}
\bf SeTformer-T & \bf49.3 & 71.5 & 53.9 & \bf44.0 & 68.3 & 47.8  & 40M  & 245  \\ \hline
%%%%%%%%%%%%%%%%%%%%%%%%%%%%%%%%%%%%%%%%%%%%%%%%%%%%%%%%%%%%%%%%%%%
 %PVT-M  & 64.1 & 302 & 44.2& 66.0& 48.2& 40.5& 63.1& 43.5 \\[-0.3ex]
Swin-S  & 48.5 & 70.3 &  53.5 &  43.3 & 67.4 & 46.5 & 69M   & 359   \\  [-0.3ex]
%ConvNeXt-S  & 48.4 &  70.8 & 53.8 & 42.6 & 67.5 & 46.8  & 69M  & 351     \\[-0.3ex]
FocalNet-S  & 49.3 & 70.7 & 54.2 &  43.8 & 68.0 & 47.4 &  72M  & 365   \\[-0.3ex]
CSWin-S &  50.0 &71.3 &54.7 &44.5 &68.4 &47.7  & 54M & 342 \\[-0.3ex]
NAT-S & 48.4 & 69.8 & 53.2 &  43.2 & 66.9 & 46.5 & 70M & 330   \\[-0.3ex]
MViTv2-S & 49.9 & 72.0 & 55.0 &  45.1 & 69.5 & 48.5 &54M & 326 \\[-0.3ex]\rowcolor{Gray}
\bf SeTformer-S & \bf51.3 & 73.1 & 55.6 & \bf45.9 & 71.2 & 49.4 & 52M & 312 \\ \hline 
%%%%%%%%%%%%%%%%%%%%%%%%%%%%%%%%%%%%%%%%%%%%
FocalNet-B & 49.8 & 70.9 & 54.6 & 44.1 & 68.2 & 47.2 & 111M & 507 \\ [-0.3ex] 
CSWin-B & 50.8 & 72.1 &55.8 &44.9 &69.1 &48.3 &97M &526 \\ [-0.3ex] 
%Swin-B  & 48.3 & 70.1 & 53.2 & 43.4 & 66.5 & 46.9  & 106M   & 498  \\[-0.3ex]
%ConvNeXt-B  &48.5 & 70.6 & 54.1 & 43.9 & 67.2 & 47.1 & 108M  & 507  \\[-0.3ex]
MViTv2-B & 51.0 & 72.7 & 56.3 & 45.7 & 69.9 &  49.6 & 71M & 392  \\[-0.3ex]\rowcolor{Gray}
\bf SeTformer-B & \bf 51.9 & 72.8 & 57.2 & \bf46.3 & 70.3 & 51.2 & 68M & 351 
%\rowcolor{Gray}
%\multicolumn{9}{c}{(b) Cascade Mask R-CNN - 3x schedule}  \\ \hline
%%%%%%%%%%%%%%%%%%%%%%%%%%%%%%%%%%%%%%%%%%%%
%ResNet-50 & 82.0 & 739 & 46.3 & 64.3& 40.1 & 61.7 & 61.7& 43.5 \\\hline
%SeTformer-M & 52.7 & 712 & 50.2 & 68.3 & 54.9 & 43.5 & 66.4 & 47.1 \\\hline
%Swin-T & 86 & 745 & 50.5 & 68.9 & 54.9 & 43.7 & 66.8 & 47.1 \\[-0.3ex]
%ConvNeXt-T & 86 & 741 &  50.4 &69.1 &54.8 &43.7 &66.5 & 47.3 \\[-0.3ex]
%CSWin-T & 80 &  757 & 52.5 & 71.5 & 57.1 & 45.3 & 68.8 & 48.9 \\[-0.3ex]
%NAT-T & 85 & 737 &51.4 &70.0 &55.9 &44.5 &67.6 & 47.9 \\[-0.3ex]
%MViTv2-T & 76 & 701 & 52.2 & 71.1 & 56.6 & 45.0 & 68.3 & 48.9 \\ [-0.3ex] \rowcolor{Gray}
%\bf SeTformer-T & 73.2 & 726 & 52.4 & 70.7 & 56.1 & 44.7 & 68.2 & 48.4\\   \hline
%Swin-S & 107 & 838 & 51.9 &70.7 &56.3 &45.0 &68.2 & 48.5 \\  [-0.3ex]
%CSWin-S & 92 &820 &53.7 &72.2 &58.4 &46.4 &69.6 &50.6 \\  [-0.3ex]
%NAT-S &108 & 809 & 52.0 &70.4 &56.3 &44.9& 68.1& 48.6 \\  [-0.3ex]
%MViTv2-S & 87 & 748 &53.2 &72.4& 58.0 &46.0 &69.6& 50.1 \\  [-0.3ex] \rowcolor{Gray}
%\bf SeTformer-S & 87 & 745 & 52.5 & 71.5 & 57.3 & 45.9 & 69.7 & 50.2 
\end{tabular}}
\vspace{-2pt}
\caption{Object detection using Mask R-CNN on the COCO. FLOPs (FL) are measured on the input resolution of 1280$\times$800. All backbones are pre-trained on ImageNet-1K.} 
\label{tab:obj}
\end{table}

%%%%%%%%%%%%%
\subsubsection{Classification on ImageNet-1K}
The ImageNet-1K \cite{deng2009imagenet} has $\sim$1.28M images in 1000 classes. Following Swin's training  setting, we utilize an AdamW \cite{kingma2014adam} for 300 iterations, with 20 for warm-up of the learning rate, followed by gradual decay, and then perform ten cool down epochs. The results of top-1 accuracy are reported in Table \ref{tab:clas}, and Figure \ref{fig1}. In our ablation study we showed that the optimal reference size for this task is 500. Raising $m$ can enhance performance until reaching a saturation point. We also set $\epsilon$ to 0.1 and $\tau$ to 0.5. 
%%%%%%%%%%%%%%%%%%%%%%%%%%%%%%%%%%%
%On the other hand, increasing the number of references while keeping the embedding dimension (i.e. p × q) constant is not significantly helpful. 
%%%%%%%%%%%%%%%%%
%%%%%%%%%%%%%%%%%%%%
%To evaluate the effectiveness of SeTformer, we compare it with different transformer- and ConvNet-based approaches on the ImageNet-1K classification dataset. ImageNet-1K  comprises 1.28M training, 50K validation, and 100K test images from 1000 classes. Following Swin's training protocol \cite{liu2021swin}, we use an AdamW optimizer \cite{kingma2014adam} for 300 epochs and 20 of which are used to warm up the learning rate, the remaining decay according to the scheduler, and then perform ten cool down epochs \cite{touvron2021training}. The batch size is 1024, the initial learning rate is set to 0.001, and the weight decay is 0.05. The results of top-1 accuracy are reported in Table \ref{tab:clas}, and Fig. \ref{fig1}. 
%%%%%%%%%%%%%%%%%%%%%%%%%%%%%%%%%%%%
SeTformers consistently outperform ConvNeXt with smaller model size, Flops, and throughput. %ConvNeXt is a CNN model based on the Swin structure, shows promising scaling behavior and outperforms Swin with similar complexity. 
%%%%%%%%%%%%%%%%%
Our mini variant exceeds Swin-T by +0.4\%, using 40\% fewer parameters (28M $\rightarrow$ 16M) and 37\% fewer Flops. Our Tiny model (83.9\%) surpasses CSWin  by +1.2\% in performance with a similar model size, resulting in a 12\% speedup (from 701/s to 785/s). It also outperforms FocalNet-T by +1.6\% while being lighter. With larger models, we achieve state-of-the-art performance with fewer parameters and lower computational costs. For example, SeTformer-B outperforms NAT-B (84.4\%) by a margin of +1.8\%, while having over 24\% and 36\% fewer Flops and parameters.  We also note that, the throughputs are measured on a V100 GPU.  
%%%%%%%%%%%%%%%%%%%%%%%%%%%%%%%%%%%%%%%%%%%%%%%%%%%%%
We also evaluate the kernel-attention models on ImageNet classification, and ensuring fairness by using the same T2T-ViT backbone for all models. With a pre-trained 24-layer model, we apply our method to 2 attention layers in the T2T module. On the ImageNet validation set, we measure top-1 accuracy and Flops of the first attention layer, which is the most resource-intensive. The results (Table \ref{kern}) highlight that our model achieves the highest accuracy at 83.17\%, while demanding 23 times fewer operations. The Performer, although the fastest, exhibits a significant -2.05\% accuracy drop compared to the vanilla transformer.

%Our small variant marginally matches its FocalNet counterpart while reducing the model size by 27\% (49.7M $\rightarrow$ 36.4M) with computation cost (8.6G $\rightarrow$ 7.2G). Our base-size model improves top-1 accuracy by 1.2\% over FacolNet-B with similar FLOPs, but again with a lower number of parameters. 

\setlength{\tabcolsep}{3pt}
\begin{table}
\centering
\small{
\begin{tabular}{@{}l|ccc@{}}  
Model & \# Param. & FLOPs & SS/MS mIoU  \\   \hline 
%{$\circ$} ResNet-101  & 86.2M & 1029G & 44.9 / - \\ \hline
\rowcolor{Gray}
SeTformer-M &  40M & 758G &  45.8 / 46.9   \\  \hline

Swin-T \cite{liu2021swin} & 60M & 946G &  44.5 / 45.8   \\ [-0.3ex]
ConvNeXt-T \cite{liu2022convnet}  & 60M & 939G  &  46.1 / 46.7\\[-0.3ex] 
%FocalNet-T \cite{yang2022focal} & 61M & 944G  &  46.5/ 47.2\\[-0.3ex]
NAT-T \cite{hassani2023neighborhood}& 58M & 934G & 47.1 / 48.4  \\[-0.3ex]
CSWin-T \cite{dong2022cswin}& 60M & 959G & 49.3 / 50.7 \\[-0.3ex] \rowcolor{Gray}
\bf SeTformer-T & 51M & 873G &  \bf50.6 / 51.4 \\ \hline

Swin-S \cite{liu2021swin}&   81M & 1040G &  47.6 / 49.5    \\[-0.3ex]  
ConvNeXt-S \cite{liu2022convnet}&  82M & 1024G &  48.7 / 49.6    \\   [-0.3ex]
%FocalNet-S  & 83M & 1035 G & 49.3 / 50.1  \\[-0.3ex]
NAT-S \cite{hassani2023neighborhood} & 82M & 1010G &  48.0 / 49.5 \\[-0.3ex]
CSWin-S \cite{dong2022cswin}& 65M & 1027G &  50.4 / 51.5 \\[-0.3ex]\rowcolor{Gray}
\bf SeTformer-S  & 59M & 986G & \bf 51.1 / 51.9  \\ \hline 

Swin-B  \cite{liu2021swin}& 121M & 1188G &  48.1 / 49.7       \\ [-0.3ex]
ConvNeXt-B \cite{liu2022convnet} & 122M & 1170G & 49.1 / 49.9    \\  [-0.3ex]
%FocalNet-B  & 124M  & 1180G &  50.2 / 51.1 \\[-0.3ex]
NAT-B \cite{hassani2023neighborhood}& 123M & 1137G & 48.5 / 49.7 \\[-0.3ex]
CSWin-B \cite{dong2022cswin}& 109M &  1222G &  51.1 / 52.2  \\[-0.3ex]\rowcolor{Gray}
\bf SeTformer-B  & 81M & 1008G & \bf 52.0 / 52.8  
\end{tabular}}
\vspace{-3pt}
\caption{Semantic segmentation on the ADE20K. FLOPs are computed based on an input resolution (2048, 512).} 
\label{tab:seg}  
\end{table}

\subsubsection{Object detection on COCO}
We perform object detection on MS-COCO \cite{lin2014microsoft}.  Models are trained on 118K training images and evaluated on 5K validation set using Mask R-CNN \cite{he2017mask}.  We pre-train the backbone on ImageNet-1k, fine-tune on COCO for 36 epochs (3$\times$schedule). We choose the number of references $m$ as 750, and set $\epsilon$ to 0.3 and $\tau$ to 0.8. %%%%%%%%%%%%%%%%%%%%%%%%%%%%%%%%%%
%On the other hand, increasing the number of references while keeping the embedding dimension (i.e. p × q) constant is not significantly helpful. 
%%%%%%%%%%%%%%%%%%%%%%%%%%%%%%%%%%%%%%%%
Table \ref{tab:obj} presents our results. SeTformer outperforms CNN (e.g., ResNet) and Transformer backbones (e.g., CSWin, NAT, MViTv2). For example, SeTformer-T (49.3/44.0) outperforms NAT-T by +1.6/+1.4 in AP$^b$/AP$^m$, while requiring less computation and having a smaller model size. When scaling up, SeTformer-B (51.9) improves over CSWin-B (50.8) by +1.1 AP$^b$, while using 28\% fewer parameters and 33\% fewer Flops. Additional results are available in the {\it Suppl. file}.

\subsubsection{Semantic segmentation on ADE20K}
We also evaluate SeTformer on ADE20K \cite{zhou2019semantic} for semantic segmentation using UperNet \cite{xiao2018unified}. Methods are trained for 160K epochs using batch size 16, following \cite{liu2021swin} (see {\it Suppl. file} for more details). Our optimal result is achieved with $m$ = 800, $\epsilon$ and $\tau$ set to 0.3 and 0.8, respectively. In Table \ref{tab:seg}, we report the mIoU and Multi-scale tested mIoU (MS mIoU) results of different methods. Our models outperform state-of-the-art methods; e.g., SeTformer-T and Setformer-S outperform CSWin counterparts by +1.3/+0.7 and +0.7/+0.4 mIoU (SS/MS), respectively, while being lighter with lower complexity.
%%%%%%%%%%%%%%%%%%%%%%%%%
%The performance of our model is evaluated using UPerNet \cite{xiao2018unified} as the base framework on ADE20K \cite{zhou2019semantic} with ImageNet pre-trained backbones. The ADE20K dataset is commonly used for semantic segmentation and contains 25K images, of which 20K are used for training, 2K for testing, and 3K for validation. There is a total of 150 semantic categories from different objects. Following the training setting of \cite{liu2021swin}, we use {\fontfamily{pcr}\selectfont{mmsegmentation}} with the following parameters: AdamW \cite{kingma2014adam} with a weight decay of 0.01 as the optimizer for 160k iterations, and the learning rate set to $6\times10^{-5}$ with a linear warm-up of 1,500 iterations at the beginning of training. Input images are resized and cropped at 512$\times$512 when training. We show the mIoU with and without multi-scale testing in Table \ref{tab:seg}. We see that our variants outperform SOTA with a large margin and fewer parameters. Specifically, SeTformer-Mini outperforms the Swin-Tiny by +0.7\% mIoU, while reducing the model size 30\% (60.0M $\rightarrow$ 41.5M), and comes very close to ConvNeXt-T (46.1 {\bf vs.} 45.8). Moreover, our tiny and small variants are +1.3\% and +1.9\% mIoU higher than the state-of-art FocalNet counterparts while significantly reducing the model size and computation cost. It is also +1.7\% and +2.9\% mIoU higher than ConvNeXt, while also being more efficient.  %In overall, these results of segmentation show the superiority of our proposed Transformer.

\subsubsection{Experiments on language modeling}
%In this section, we investigate SeTformer's efficacy on the GLUE (QQP, SST-2, MNLI) \cite{wang2018glue} benchmark. To ensure fairness, all transformers, including SeTformer, are pre-trained for 50K iterations on the WikiText-103 \cite{merity2016pointer}. Subsequently, we fine-tune the models on downstream tasks following the RoBERTa \cite{liu2019roberta}. As shown in Table \ref{nlp}, SeTformer outperforms the baseline \cite{liu2019roberta}, achieving competitive or superior results across downstream datasets compared to other efficient transformers. While our model matches the performance of Longformer \cite{beltagy2020longformer} on MNLI, Longformer exhibits a slower runtime due to its O(nw) complexity. Our model outperforms other competing kernel-based methods by a significant margin. The GLUE benchmark is a diverse set of language tasks that hinge on context and word relationships. SeTformer's unique attention mechanism based on optimal transport and kernel learning allows it to effectively model these relationships, potentially leading to improved performance.
%%%%%%%%%%%%%%%%%%%%%%%%%%%%%%%%%%%%%%%%%%%%%%%%%%%%%%%%%%%
In this section, we investigate SeTformer's efficacy on the GLUE benchmark \cite{wang2018glue} across QQP, SST-2, and MNLI tasks. To ensure fairness, all transformers, including SeTformer, are pre-trained for 50K iterations on the WikiText-103 \cite{merity2016pointer}. We fine-tune the models using training parameters from RoBERTa \cite{liu2019roberta}. By comparing SeTformer with other scalable transformers in Table \ref{nlp}, we observe that SeTformer consistently outperforms the baseline \cite{liu2019roberta}. It achieves competitive or superior performance across various downstream datasets in comparison to other efficient transformers. The GLUE benchmark is a diverse set of language tasks that hinge on context and word relationships. SeTformer's unique attention mechanism based on optimal transport 
 allows it to effectively model these relationships, potentially leading to improved performance.

 %Our goal in this section is to demonstrate the effectiveness of SeTformer's unique features in improving performance on the GLUE benchmark, including QQP (Quora Question Pairs), SST-2 (Stanford Sentiment Treebank), and MNLI (MultiNLI). We use our mini SeTformer model with 64 Nystr\"om filters. The input features for our model consists of word vectors with a dimension of 768, as provided by the HuggingFace implementation of the transformer BERT \cite{devlin-etal-2019-bert}. To ensure a fair evaluation, we pre-train all the transformers, including SeTformer, for an equal number of iterations (50K) on the WikiText-103 dataset \cite{}. Subsequently, we adopt the fine-tuning procedure employed by RoBERTa \cite{}, to fine-tune these models on the downstream tasks. The results are reported in Table \ref{nlp}.

\setlength{\tabcolsep}{3pt}
\begin{table}
\centering
\small{
\begin{tabular}{@{}l|cccc@{}}  
Model & QQP & SST-2 & MNLI     \\   \hline
%& Acc &  Test (im/s) & $\text{AP}^{\text{box}}$ & Train(iter/s) & Test(im/s)  \\   \hline \hline
%BERT-base   &  91.9 & 89.6& 92.7    \\  [-0.4ex]
Roberta-base \cite{liu2019roberta} & 88.41 & \bf92.31 & 79.15   \\  \hline 
Performer \cite{dosovitskiy2020image}& 69.92 &50.91 &35.37  \\ [-0.3ex] 
Reformer \cite{kitaev2020reformer} & 63.18 &50.92 &35.47 \\ [-0.3ex] 
Linear Trans. \cite{katharopoulos2020transformers} & 74.85 & 84.63 & 66.56 \\[-0.3ex] 
Longformer \cite{beltagy2020longformer}& 85.51 &88.65 &77.22  \\[-0.3ex] 
%FAVOR# \cite{} &   88.43 &  90.85 &  78.64  \\[-0.3ex] 
FAVOR++ \cite{likhosherstov2022chefs} & 88.43 &  92.23 & 78.91 \\ [-0.3ex] \rowcolor{Gray}
%KL-Trans \cite{chowdhury2022learning} & 75.28 &76.49 &57.6    \\  [-0.3ex]   
\bf SeTformer & \bf 88.76 &  92.28 & \bf 79.16 \\ [-0.3ex] 
\end{tabular}}
\vspace{-2pt}
\caption{classification results on GLUE for different kernel-based transformers using \cite{liu2019roberta} as baseline.} 
\label{nlp}
\end{table}
%\vspace{-15pt}
%The GLUE benchmark includes various natural language understanding tasks, where understanding context and relationships between words is crucial. SeTformer's unique attention mechanism based on optimal transport and kernel learning allows it to effectively model these relationships, potentially leading to improved performance on tasks that require understanding and reasoning about textual data. Additionally, SeTformer's computational efficiency could also contribute to faster training and inference times, which is beneficial when dealing with large datasets like GLUE; in contrast to other transformer, which would have stored a 1000$\times$1000 attention matrix, our attention score with a reference of size 64 is only 1000$\times$64. 

\setlength{\tabcolsep}{1.5pt}
\begin{table}
\centering
\small{
\begin{tabular}{@{}l|ccc@{}}   
\multirow{2}{*}{Model / datasets} & \multicolumn{1}{c|}{ImageNet}  & \multicolumn{2}{c}{COCO}     \\   \cline{2-4}
& \multicolumn{1}{c|}{Top/1}  & $\text{AP}^{\text{b}}$ & Train(iter/s) \\   \hline 
1) no pos. &  \multicolumn{1}{c|}{83.78}  & 49.63 & 3.2   \\  [-0.3ex]
2) APE  \cite{dosovitskiy2020image}  & \multicolumn{1}{c|}{83.96}  & 50.41 & 2.9  \\ [-0.3ex]
3) RPE  \cite{liu2021swin}  & \multicolumn{1}{c|}{84.35}  & 50.89 & 1.5 \\  [-0.3ex] \rowcolor{Gray}
4) LPE (ours) & \multicolumn{1}{c|}{84.72}  & 51.32 & 3.1  \\  \hline
\multirow{2}{*}{Model / datasets} &  \multicolumn{3}{c}{GLUE}  \\ \cline{2-4}
& QQP & SST-2 & MNLI \\ \hline
{5) dot product instead of OT} & 87.85& 89.26 & 78.03  \\ [-0.3ex]
{6) Sliced-OT \cite{kolouri2020wasserstein}} & 86.74 & 88.96 & 76.51 \\ [-0.3ex] 
%Sliced-Was  + pos emb. \cite{devlin-etal-2019-bert} & 86.55 &87.39&74.28\\
{7) $\mathcal{K}$ }& 88.92 & 91.73 & 78.60 \\ [-0.3ex]
{8) $\mathcal{K}$ (w/o position encoding) }& 86.32 & 88.25 & 77.38 \\ [-0.3ex] \rowcolor{Gray}
%{9) $\mathcal{K}_y$ (w/o pos. enc.) }& 88.32 & 89.77 & 77.58 \\  [-0.3ex] 
{9) $\mathcal{K}_y$}  & \bf 89.36 & \bf 93.54 & \bf 79.81  
\end{tabular}}
\vspace{-2pt}
\caption{Ablation study. Rows [1-4] explore positional bias; Rows [5-6] examine the effect of using OT; Rows[7-8] analyze the reference impact in the SeTformer-S. } 
\label{tab:ab}
\end{table}
%no pos.: without incorporating positional encoding; APE: absolute position encoding; RPE: relative position encoding; ours is the default setting of our model detailed in Section \ref{sec:pos}.

\setlength{\tabcolsep}{3pt}
\begin{table}
\centering
\small{
\begin{tabular}{@{}l|lclc@{}}  
{Model} & No. of ref.  ($m$) & ImageNet  & {COCO} & ADE20K    \\ \hline
%& No. of reference &  IN & COCO & SST-2   \\     [-0.4ex]
\multirow{5}{*}{SeTformer-T}  &  100 &   79.81 & 45.76 & 46.93  \\     [-0.5ex]
%12) - (w/o pos.) & &     \\     [-0.4ex]  
  & 300 & 82.54 & 46.59 & 48.21    \\    [-0.5ex]  
%14) ours (w/o pos.) &  &      \\    [-0.4ex]  
 & 500 &  \bf 83.91 & 48.55 & 49.11  \\    [-0.5ex]  
%3) ours (w/o pos.) & &   \\       [-0.4ex]  
 & 750 / 800 & 83.25 & {\bf{49.34}} / 49.37 &\bf50.62  \\     [-0.5ex]  
 & 1000 &   81.74 & 48.19 & 50.08   
\end{tabular}}
\vspace{-5pt}
\caption{Number of $m$ vs. performance:  performance improves with increasing $m$, but saturates at high values.}  
\label{no.of.ref}
\end{table}

%%%%%%%%%%%%%%%%%%%%%%%%%%%
\setlength{\tabcolsep}{2pt}
\begin{table}
\centering
\small{
\begin{tabular}{@{}l|cc|ccc@{}}  
\multirow{2}{*}{Model} &  \multicolumn{2}{c}{ImageNet-1k} & \multicolumn{3}{c}{COCO}     \\   \cline{2-6}
& Acc. &  Test (im/s) & $\text{AP}^{\text{box}}$ & Train(iter/s) & GFLOPs \\   \hline 
%1) SWin-B &  83.3 & 278 & 48.3 & 3.1    \\  [-0.4ex]
 MViTv2-B &  84.4 & 253 & 51.0 & 2.1 & 392   \\   \hline \rowcolor{Gray}
 SeTformer-S & 84.7 & 451 & 51.3 & 3.1  & 245\\  [-0.5ex]  
 SeTformer-B & 86.2 & 298 & 51.9 & 2.4 & 371
\end{tabular}}
\vspace{-2pt}
\caption{Runtime comparison on ImageNet-1K and COCO. }  
\label{runtime}
\end{table}

\subsection{Ablation studies}
{\bf Positional encoding comparison:} In this section, we evaluate the important components of the SeTformer using ImageNet, COCO, and GLUE datasets. Table \ref{tab:ab}, Rows [1-4] ablates the different position embedding techniques. We find: (i) Absolute positions (2) only slightly improve over no position (1), as pooling operators already encode position. (ii) Relative positions (3) boost performance by introducing shift-invariance. Notably, our linear positional information proves remarkably efficient, particularly in COCO, where it accelerates training by 2.1$\times$ versus relative positions. {\bf Replacing OT with other variants:} Rows [5-6] utilize dot product \cite{vaswani2017attention} and sliced-OT instead of OT. The results are less favorable compared to our model. {\bf Effect of using references:} Rows [7-8] explore the impact of references. (7) employs $\mathcal{K}$ without the reference concept, while (8) indicates using $\mathcal{K}$ without adding positional encoding. Comparing Rows (9) with (7), the presence of references enhances performance.  
%%%%%%%%%%%%%%%%%%%%%%%%%%%%%%%%%%%%%%%%%%%%%%%%%%%%%%%%%%%
We analyze the model's sensitivity to the number of elements $m$ in Table \ref{no.of.ref}. For example, in the COCO scenario, using m = 800 yields slightly improved results (49.37), while m = 750 also provides good performance (49.34) and it has the advantage of being lighter. While increasing m can enhance performance, it becomes less impactful for very large values.
 %%%%%%%%%%%%%%%%%%%%%%%%%%%%%%%%%%%%%%%%% 
%%%%
{\bf Runtime comparison:} The runtime comparison between SeTformer, NAT, and Swin is presented in Table \ref{runtime}. SeTformer-S surpasses MViT-B in both ImageNet (+0.3\%) and COCO (+1.2\%) datasets, while having higher throughput (451 im/s vs. 253 im/s) on ImageNet and faster training (3.1 iter/s vs. 2.1 iter/s) on COCO. SeTformer-B is slightly slower but remarkably more accurate (+1.8\% on ImageNet-1K and +0.9\% on COCO. 

\subsection{Robustness to Sequence Length Variation}
\begin{figure}
    \centering
    \includegraphics[width=6.5cm]{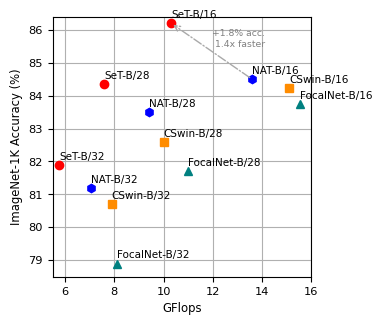}
    \caption{{\small Performance of SeTformer-B and SOTA DSPA counterparts scaled with different patch sizes. Results show that for shorter sequence lengths (larger patches), performance is similar across most models, but SeT-B and NAT-B lead in computational efficiency with lower FLOPs. }}
       \label{fig}
\end{figure}

We evaluated the performance of various models using different patch sizes: /32, /28 and /16. For fairness, we used the Base-size version of all models. The objective is to show the trade-offs between computational cost and accuracy as a function of sequence length - smaller patches mean longer sequences. Fig.\ref{fig} indicates that while smaller patches (longer sequences) yield higher accuracy due to finer details in the input data, they also significantly increase the computation in terms of FLOPs. Our SeT-B consistently outperformed the baselines across varying sequence lengths, achieving higher accuracy with lower FLOPs. FocalNet-B exhibits a considerable drop in performance with shorter sequences. This suggests an inefficiency in handling shorter sequence lengths while performing better on longer sequence lengths. 

\vspace{-5pt}

\section{Conclusion}
We proposed SeTformer, an efficient transformer based on Self-optimal Transport (SeT).  Our SeTformer benefits from the properties of kernel methods and optimal transport, effectively captures complex relationships and ensures that attention scores are both non-negative and follow a reweighting scheme similar to the softmax mechanism. By leveraging this unique structure, we have created a series of SeTformer-X architectures that consistently surpass existing transformers across vision and language domains.  Our work can provide another perspective on attention modulation to explore better content-based interactions, which can benefit visual recognition models. 

\section{Acknowledgments}
In this paper, P. Shamsolmoali and M. Zareapoor have made equal contributions. This work was partly supported by the Wallenberg Artificial Intelligence, Autonomous Systems and Software Program (WASP), funded by the Knut and Alice Wallenberg Foundation, and the Swedish Research Council grant 2022-04266.

%The proposed SeT is a novel self-attention model for embedding feature vectors and then aggregating them for visual modeling, which consists of optimal transport and kernel methods. To capture the effective dependencies (from short to long ranges), we learned our mechanism in the RKHS. Then we used optimal transport for aggregating the input feature vectors based on the reference set. Moreover, we used a kernel approximation method to keep the computational complexity linear with no additional structure to improve the scalability. With the SeT, we built a series of transformer architectures (SeTformer-X), which can significantly outperform state-of-art transformers, e.g., Swin and Focal, but with a lower computation complexity on the tasks of image classification, object detection, and segmentation. Our work can provide another perspective on attention modulation to explore better content-based interactions, which can benefit visual recognition models. 

\bigskip

\bibliography{aaai24}

\newpage

\definecolor{Gray}{gray}{0.8}

\subsection{Experiment details on language modeling}
To ensure a fair evaluation, we begin by pre-training all efficient transformers for an equal number of 50,000 iterations on WikiText103 (Merity et al., 2017). We then use the fine-tuning protocol outlined in RoBERTa \cite{liu2019roberta} to fine-tune these methods on Amazon dataset. As shown in Table \ref{nlp}, Setformer consistently outperforms the baseline established by \citet{liu2019roberta}, demonstrating superior or competitive results when compared with other efficient transformers. Although Longformer \cite{beltagy2020longformer} achieves comparable results, its computational complexity of $\mathcal{O}(nk)$, where k signifies the window size, results in slower execution and greater memory consumption than Setformer. Other methods, which rely on kernel functions, exhibit a significant performance gap compared to our model. This proves the efficacy of our SeTformer in comparison to alternative efficient transformer variants.

\begin{table}
    \centering
      \small{
    \begin{tabular}{cc}\\
& Range of values \\ \hline
Entropic regularization $\epsilon$ & [0.3; 0.5; 1.0]\\ \hline
bandwidth in position encoding $\tau$ & [0.1, 0.3; 0.5; 0.6]\\ \hline
Learning rate &  [0.1; 0.01; 0.001] \\\hline
 \end{tabular}}
 \caption{Hyperparameter range for language modeling task.}
\label{param.}
\end{table}

\subsection{Experiment details on COCO Object detection}
For our object detection experiments, two popular object detection models are used: Mask R-CNN and Cascade Mask R-CNN, as implemented in mmdetection \cite{chen2019mmdetection}. Our training protocol aligns with that of Swin, involving multi-scale training (resizing the shorter side from 480 to 800, while keeping the longer side below 1333) and a 3x training schedule (36 epochs). We employ the AdamW optimizer, setting the learning rate to 0.0001, incorporating a weight decay of 0.05, and operating on a batch size of 16. The learning rate reduces at the 8th and 11th epochs with decay rates of 0.1. Our model's hyperparameters consist of the number of reference $m$, the entropic regularization $\epsilon$, and the position encoding $\tau$.  The exploration ranges for these parameters are detailed in Table \ref{param.}, and specific results can be found in Table \ref{acc.}. Our best results are achieved by $\epsilon$ set to 0.3 and $\tau$ to 0.8. The result of the Mask R-CNN are presented in the main paper, and the Cascade Mask R-CNN results can be found in Table \ref{cas}. 
%An illustration of the proposed FAKOT is presented in Fig. \ref{matrix}.  Given $x$ as feature vectors with $n$ elements, which is extracted from input images, we  perform a spherical K-mean algorithm to obtain reference set $y$, where $y_1, ..., y_m$.  Indeed, using the reference $y$ we reduced the attention matrix from $n\times n$ to $n\times m$, where, $m$ is the elements of the reference. As illustrated in Fig. \ref{matrix}, if the paired input and reference $(x_i, y_j)$ are matched, the optimal transport plan $T$ satisfies that the embedded feature can be used for aggregation, and form $Att_y(x)$. 

%\setlength{\tabcolsep}{3pt}
\begin{table}
   \centering
    \small{
\begin{tabular}{lcc}\\ 
& & acc. (\%) \\ \hline
Vanilla Transformer \cite{liu2019roberta} & $\mathcal{O}(n^2)$ & 75.79 \\\hline
RFA \cite{peng2021random} & $\mathcal{O}(n^2)$ & 69.58  \\
Reformer \cite{kitaev2020reformer} & $\mathcal{O}(Llog L)$ & 65.43  \\
Longformer \cite{beltagy2020longformer} & $\mathcal{O}(LK)$ & 75.71 \\
Linformer \cite{wang2020linformer} & $\mathcal{O}(L)$ &  63.81\\
LinearElu \cite{likhosherstov2022chefs} & $\mathcal{O}(L)$ &  73.89\\
Performer \cite{choromanski2020rethinking} & $\mathcal{O}(L)$ &  65.89\\
SeTformer with m = 15 & $\mathcal{O}(nm)$ & 74.63 \\ \rowcolor{Gray}
SeTformer with m = 64 &  & 76.12 \\
SeTformer with m = 120 &  & 75.38 \\ \hline
    \end{tabular}}
    \caption{Classification accuracy on Amazon. Here, n represents the sequence length, K denotes the size of a window, m is the number of reference in our model.}
    \label{nlp}
\end{table}

\begin{table}
    \centering
      \small{
    \begin{tabular}{c c} \\
    & Range of values \\\hline
Entropic regularization $\epsilon$ & [0.001, 0.01, 0.1, 0.3, 0.5] \\
Position encoding $\tau$ & [0.5; 0.6; 0.7; 0.8; 0.9; 1.0] \\
Learning rate &  [0.01; 0.001; 0.0001] \\\hline
    \end{tabular}}
    \caption{Hyperparameter range for COCO and ADE20K. }
    \label{tab:my_label}
\end{table}

\setlength{\tabcolsep}{3pt}
\begin{table}[h!]
\centering
\small{
\begin{tabular}[t]{llcc}  
Dataset & \# of classes & Training size & Test size    \\   \hline  
ImageNet-1K & 1K & 1.2M & 100K  \\     
Fashion-MNIST   & 10& 60K &10K \\  
I-naturalist2021   & 10K &2.7M& 500K  \\  
Places365  & 365& 1.8M &328K \\   \hline
\end{tabular}}
\caption{Details of the classification datasets for Figure \ref{compare}.} \label{tab1}
\end{table}

\definecolor{Gray}{gray}{0.8}
%>{\color{Gray}}c >{\color{Gray}}c

\setlength{\tabcolsep}{6pt}
\begin{table*}
\centering
\small{
%{\centering (Throughput also are measured with the same resolution using \cite{}, and a V100 GPU.)}
\begin{tabular}[t]{@{}l| r c |c c cc c c   @{}}  
%\rowcolor{Gray}
 \multicolumn{2}{c}{ }& \multicolumn{6}{c}{Cascade Mask R-CNN - 3x schedule}  \\ \hline
Model & Param. & Flops & $\text{AP}^b$ &$\text{AP}_{50}^b$ & $\text{AP}_{75}^b$ &$\text{AP}^m$  &$\text{AP}_{50}^m$  & $\text{AP}_{75}^m$ \\ \hline  
%& (M) & (G) & \\ 
%\multicolumn{9}{c}{Mask R-CNN, ~ $3\times$ schedule}  \\ \hline  
%%%%%%%%%%%%%%%%%%%%%%%%%%%%%%%%%%%%%%%%%%%%
%ResNet-50 & 82.0 & 739 & 46.3 & 64.3& 40.1 & 61.7 & 61.7& 43.5 \\\hline
SeTformer-M & 57M & 681G & 50.4 & 68.7 & 54.9 & 43.5 & 66.7 & 47.2 \\\hline

Swin-T & 86M  & 745G & 50.5 & 69.3 & 54.9 & 43.7 & 66.6 & 47.1  \\[-0.3ex]
ConvNeXt-T & 86M & 741G &  50.4 &69.1 &54.8 &43.7 &66.5 & 47.3 \\[-0.3ex]
CSWin-T & 80M &  757G & 52.5 & 71.5 & 57.1 & 45.3 & 68.8 & 48.9 \\[-0.3ex]
NAT-T & 85M & 737G &51.4 &70.0 &55.9 &44.5 &67.6 & 47.9 \\[-0.3ex]
MViTv2-T & 76M & 701G & 52.2 & 71.1 & 56.6 & 45.0 & 68.3 & 48.9 \\ [-0.3ex] \rowcolor{Gray}

\bf SeTformer-T & 72M & 701G & \bf 52.9 & 72.1 & 57.8 & \bf46.2 & 68.9 & 49.3\\   \hline

Swin-S & 107M & 838G & 51.8 &70.4 &56.3 &44.7 &67.9 & 48.5 \\  [-0.3ex]
ConvNexT-S &  108M &827G &51.9 &70.8 &56.5 &45.0& 68.4& 49.1  \\
CSWin-S & 92M & 820G & 53.7 & 72.2 &58.4 &46.4 &69.6 &50.6 \\  [-0.3ex]
NAT-S &108M & 809G & 52.0 &70.4 &56.3 &44.9& 68.1& 48.6 \\  [-0.3ex]
MViTv2-S & 87M & 748G & 53.2 &72.4& 58.0 &46.0 &69.6& 50.1 \\  [-0.3ex] \rowcolor{Gray}
\bf SeTformer-S & 85M & 746G & \bf 54.3 & 72.5 & 59.1 & \bf46.8 & 69.7 & 51.4 
\end{tabular}}
%\vspace{-2pt}
\caption{Object detection performance evaluation using Cascade Mask R-CNN on COCO Dataset. FLOPS are based on an input resolution of (1280, 800). While Cascade Mask R-CNN is a more powerful model compared to the Mask R-CNN, our SeTformer consistently outperforms the competing models across various configurations.} 
\label{cas}
\end{table*}

\definecolor{Gray}{gray}{0.8}
\begin{table*}[t]
    \centering
    \small{
    \begin{tabular}{lcccc} 
  &   & ImageNet-1k  & COCO & ADE20K \\ \hline
Model & No. of reference & acc. & $AP^b$ &  mIoU \\ \hline
SeTformer-S & \multirow{2}{*}{ m=100} &  81.36 &  47.21 & 46.39 \\
SeTformer-S  (with pos. enc.)&  &  82.64 & 49.32 &  47.85 \\\hline
SeTformer-S & \multirow{2}{*}{ m=500} &  82.26 & 48.54 &   47.63 \\
SeTformer-S (with pos. enc.)&  & \bf84.71  &  50.25 &  49.36 \\\hline
SeTformer-S & \multirow{2}{*}{ m=750} &  83.15 &  49.76 &  49.22  \\
SeTformer-S (with pos. enc.)&  & 84.25  & \bf51.30 &  51.08 \\  \hline
SeTformer-S & \multirow{2}{*}{ m=800} &  83.27 &  49.82 &  49.23  \\
SeTformer-S (with pos. enc.)&  & 84.25  & 51.31 &  \bf51.10 \\  \hline
SeTformer-S & \multirow{2}{*}{ m=1000} & 82.85 & 49.78 &  49.31\\
SeTformer-S (with pos. enc.)&  & 83.78  & 51.23 &  50.82\\ \hline
    \end{tabular}}
    \caption{Experiments based on number of reference $m$. We here consider the impact of of the number of references ($m$) on the performance of our model, each learned through K-means clustering. The investigation is carried out with and without the integration of positional embeddings. We observe that our model's performance is moderately influenced by the number of $m$. Our experiments highlight that while augmenting the number of $m$ can lead to improved results, but becomes less significant as $m$ grows excessively. Indeed, increasing $m$ tends to enhance the performance of our model, but more values for $m$ is not significantly helpful. We also observe that the inclusion of positional information significantly influences this task, leading to  performance improvement compared to the non-encoded variant.}
    \label{acc.}
\end{table*}

\definecolor{Gray}{gray}{0.8}
\begin{table*}[!h]
    \centering
    \begin{tabular}{lcccc|cccc} 
  &   \multicolumn{4}{c}{Cascade mask R-CNN on COCO} &   \multicolumn{4}{c}{UperNet on ADE20K} \\ \hline
& Param. & FLOPs & FPS & $\text{AP}^{box/mask}$ &  Param. & FLOPs & FPS & mIoU \\ \hline
Swin-T  &  86M &745G &15.3 &50.5/43.7 & 60M &946G &18.5& 44.5  \\
CSWin-T     &   80M &757G &14.2& 52.5/45.3 &60M& 959G &17.3 &49.3 \\   
%MViTv2-T & 76M&  701G& 11.6 & 52.2/45.0 &  \\ 
SeTformer-T  & 72M & 701G & 15.8&  52.9/46.2& 51M &873G & 18.7 & 50.6    \\\hline
Swin-S  &  107M &838G &12.0 &51.8/44.7& 81M &1038G& 15.2& 47.6  \\
CSWin-S    & 91M &820G &11.7 &53.7/46.4 & 65M &1027G &15.6 &50.4  \\
SeTformer-S    &   85M & 746G & 12.3 & 54.3/46.8 & 59M &986G &16.0& 51.1 \\ \hline
    \end{tabular}
    \caption{Throughput (FPS) compared to Swin, and CSWin on downstream tasks. Our tiny model (SeTformer-T) achieves better performance than Swin-S on COCO, with an improvement of +1.1\% in box AP and +1.5\% in mask AP, while maintaining significantly higher throughput (15.8 vs. 12) and using  32\% fewer parameters.  }
    \label{infe}
   % \vspace{-2cm}
\end{table*}

\subsection{Experiment details on ADE20K semantic segmentation}
We evaluate our proposed model for the task of semantic segmentation using UperNet \cite{xiao2018unified} framework, implemented in mmsegmentation \cite{contributors2020mmsegmentation}. We follow a setup similar to Swin, employing the AdamW optimizer with an initial learning rate of 6e-5, weight decay of 0.01, and a batch size of 16 for 160K iterations. We incorporate learning rate warm-up for the first 1500 iterations, followed by linear decay. The input size 512$\times$512 is used for training all models. During testing, results are reported for both single scale and multi-scale tests (SS/MS mIoU). Our model's hyperparameters consist of the number of reference $m$, the entropic regularization $\epsilon$, and the position encoding $\tau$.  The exploration ranges for these parameters are detailed in Table \ref{param.}, and specific results can be found in Table \ref{acc.}(ADE20K). We achieve our optimal results with $\epsilon=0.3$ and $\tau=0.8$. It's important to note that extremely small values of $\epsilon$ (e.g., 0.01) can lead to null elements in a matrix product, hence, the ideal choices for $\epsilon$ in this context is 0.3 to 0.6.

\begin{figure*}[!t]
\centering
 \includegraphics[width=15.5cm]{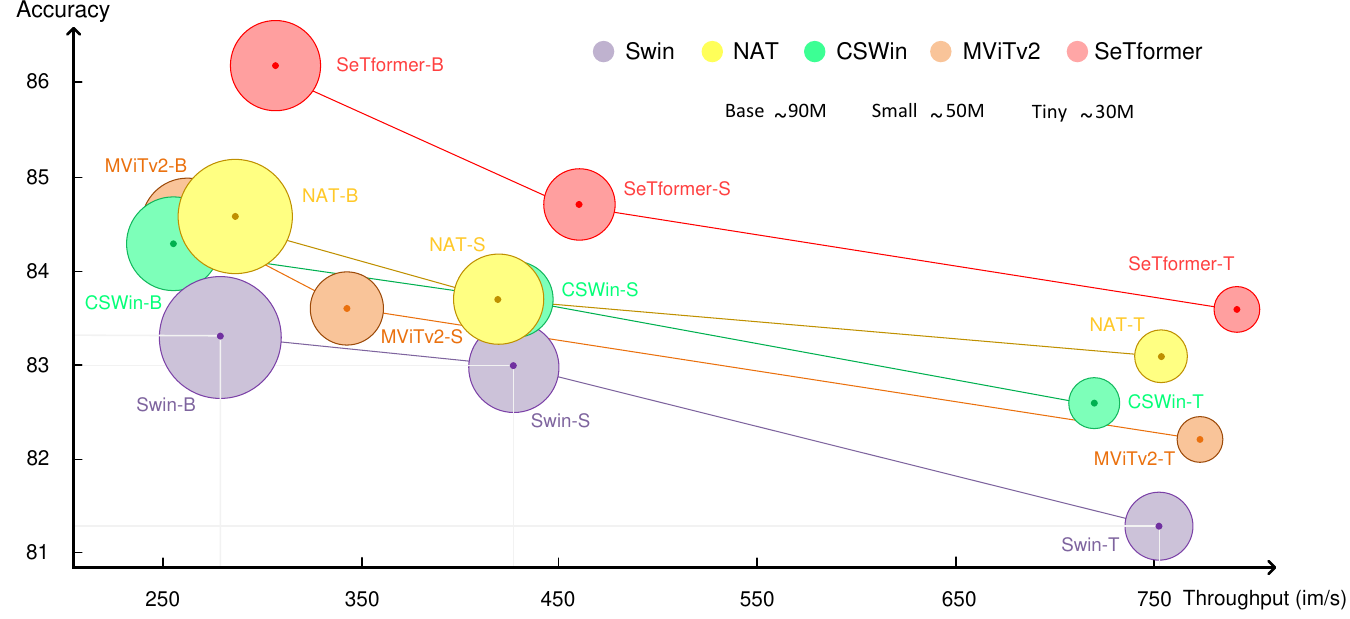}
 \caption{Throughput vs. accuracy on ImageNet-1k. Our model surpasses the baseline models in terms of top-1 accuracy while maintaining higher throughput. } \label{throu}
%   \vspace{-2cm}
\end{figure*}

\subsection{Remarks on model efficiency}
We provide insights into the efficiency of the CSWin, Swin, and SeTformer models. We evaluate the inference speed across various downstream tasks, presenting the throughput (FPS) for the Cascade Mask R-CNN applied to the COCO dataset and UperNet for semantic segmentation on the ADE20K dataset. As can be seen in the Table \ref{infe}, our SeTformer shows superior speed compared to other baseline methods. Swin displays a slightly higher inference speed compared to CSWin (less than 10\% faster). On the COCO dataset, SeTformer-S outperforms Swin-S/CSWin-S by +2.5\%/+0.6\% in box AP and +2.1\%/+0.4\% in mask AP, while maintaining a higher inference speed (12.0/11.7 FPS vs. 12.3 FPS). Notably, SeTformer-T surpasses Swin-S in both box AP (+1.1\%) and mask AP (+1.5\%), all while achieving significantly faster inference speed (15.8 FPS vs. 12 FPS), highlighting the effective accuracy/FPS balance achieved by our SeTformer model. Figure \ref{throu} shows the top/1 accuracy vs. throughput of the recent efficient transformers. Our model surpasses the baseline models in terms of top-1 accuracy while maintaining higher throughput.  Specifically, our model achieves top-1 accuracy of 83.9\% and 84.7\% using tiny and base model sizes, respectively. Moreover, our model's throughput is comparable to that of MViTv2-T and approximately 40\% higher than MViTv2-S.

\end{document}